\documentclass[runningheads]{llncs}
\usepackage{times}
\usepackage{epsfig}
\usepackage{graphicx}
\usepackage{amsmath}
\usepackage{amssymb}
\usepackage{subcaption}
\usepackage{booktabs}
\usepackage{amsfonts}
\usepackage{array}
\usepackage{textcomp}
\usepackage{multirow}
\usepackage{makecell}
\usepackage{diagbox}
\usepackage{threeparttable}
\usepackage{mathtools}
\usepackage{eqparbox}
\usepackage{bm}
\usepackage{bbding}
\usepackage{comment}
\usepackage{color}
\usepackage{bbding}
\usepackage{pifont}
\usepackage{wasysym}
\usepackage{cite}
\usepackage{stfloats}
\usepackage{float}

\DeclareMathOperator{\y}{\mathbf{a}}

\DeclareMathOperator{\C}{\mathbf{C}}
\DeclareMathOperator{\N}{\mathcal{N}}
\DeclareMathOperator{\I}{\mathbf{I}}
\DeclareMathOperator{\x}{\mathbf{x}}

\newcommand{\tabincell}[2]{\begin{tabular}{@{}#1@{}}#2\end{tabular}}

\begin{document}
%
\title{Bootstrap Diffusion Model Curve Estimation for High Resolution Low-Light Image Enhancement
\thanks{J. Huang and Y. Liu—Contributed
equally to this work. \\This work is supported by Key-Area Research and Development Program of Guangdong Province (2019B010155003), the Joint Lab of CAS-HK, and Shenzhen Science and Technology Innovation Commission (JCYJ20200109114835623, JSGG20220831105002004).}}
\titlerunning{Bootstrap Diffusion Model Curve Estimation for High Resolution for LLIE}
%

\author{Jiancheng Huang\inst{1,2} \and
Yifan Liu\inst{1} \and
Shifeng Chen\inst{1,2}(\Envelope)}
\authorrunning{J. Huang et al.}
%
\institute{ShenZhen Key Lab of Computer Vision and Pattern Recognition, Shenzhen Institute of Advanced Technology, Chinese Academy of Sciences, Shenzhen, China \\\email{\{jc.huang,yf.liu2,shifeng.chen\}@siat.ac.cn} \and
University of Chinese Academy of Sciences, Beijing, China
}

\maketitle              
\begin{abstract}
Learning-based methods have attracted a lot of research attention and led to significant improvements in low-light image enhancement. However, most of them still suffer from two main problems: expensive computational cost in high resolution images and unsatisfactory performance in simultaneous enhancement and denoising. To address these problems, we propose BDCE, a bootstrap diffusion model that exploits the learning of the distribution of the curve parameters instead of the normal-light image itself. 
Specifically, we adopt the curve estimation method to handle the high-resolution images, where the curve parameters are estimated by our bootstrap diffusion model. In addition, a denoise module is applied in each iteration of curve adjustment to denoise the intermediate enhanced result of each iteration. 
We evaluate BDCE on commonly used benchmark datasets, and extensive experiments show that it achieves state-of-the-art qualitative and quantitative performance.

\end{abstract}
\section{Introduction}\label{intro}

Low-light image enhancement (LLIE) is a very important and meaningful task in computer vision. Images captured in environments with insufficient lighting often exhibit numerous issues, including diminished contrast, dark colors, low visibility, etc. Therefore, LLIE is often used to process these poor quality images, and the processed images can also be better suited for other downstream tasks~\cite{huang2023ifast}.

The traditional LLIE method is mainly implemented based on histogram equalization~\cite{ibrahim2007brightness,abdullah2007dynamic} and Retinex model~\cite{wang2013naturalness,park2017low,li2018structure}. However, these methods still do not do a good job of detail and accurate color restoration. In the past few years, with the continuous development of deep learning, there are more and more LLIE methods based on deep learning~\cite{lore2017llnet,jiang2021enlightengan,zhang2021star,yang2020fidelity,Wang_2019_CVPR,guo2020zero,liu2021retinex,zhang2022deep,wu2022uretinex,yang2022adaint,xu2022snr,ma2022toward,kim2021representative,Wang_2022_ECCV,Yang_2022_ECCV,10268075}. Compared with traditional methods, these methods achieve better visual effects and are more robust.

Despite the progress of existing methods, two major problems remain: 1) The first problem is that LLIE on high resolution images is computationally too expensive. 2) The second problem is that simultaneous enhancement and denoising is unsatisfactory. Solving both problems is extremely challenging.

\begin{figure*}[t]
  \small
  \centering
  \setlength{\abovecaptionskip}{-0.3cm}
  \captionsetup[subfigure]{labelformat=empty}  
  \begin{subfigure}{0.16\linewidth}
    \includegraphics[width=1.98cm]{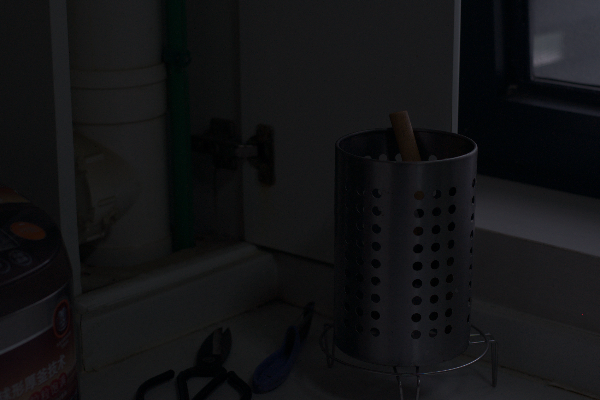}
    \centering
    \centerline{Input}
  \end{subfigure}
  \begin{subfigure}{0.16\linewidth}
    \includegraphics[width=1.98cm]{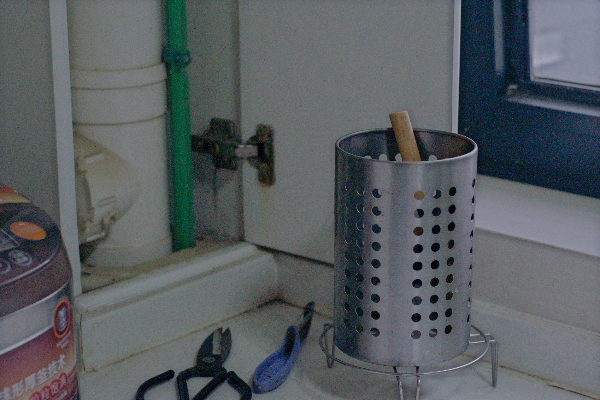}
    \centering
    \centerline{${\mathrm{naive}}$}
  \end{subfigure}
  \begin{subfigure}{0.16\linewidth}
    \includegraphics[width=1.98cm]{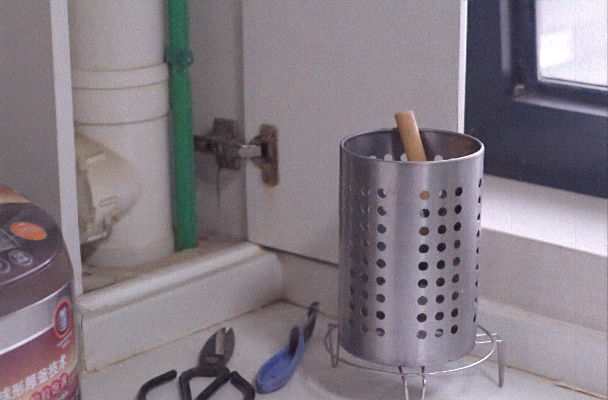}
     \centering
     \centerline{${\mathrm{w/o\ denoise}}$}
  \end{subfigure}
  \begin{subfigure}{0.16\linewidth}
    \includegraphics[width=1.98cm]{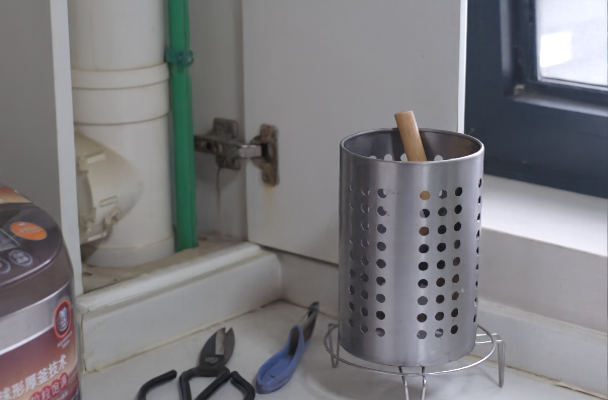}
    \centering
    \centerline{${\mathrm{w/o\ diff}}$}
  \end{subfigure}
  \begin{subfigure}{0.16\linewidth}
    \includegraphics[width=1.98cm]{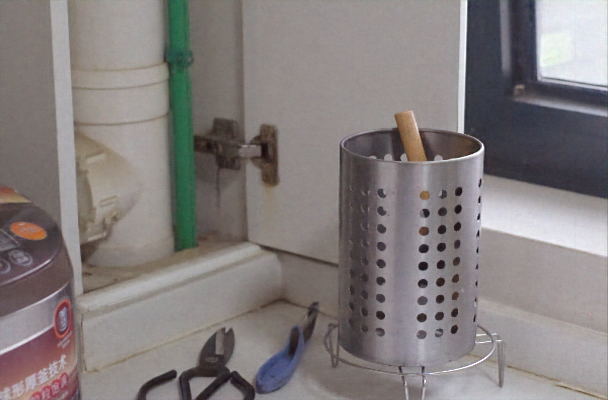}
    \centering
    \centerline{${\mathrm{w/o\ self}}$}
  \end{subfigure}
  \begin{subfigure}{0.16\linewidth}
    \includegraphics[width=1.98cm]{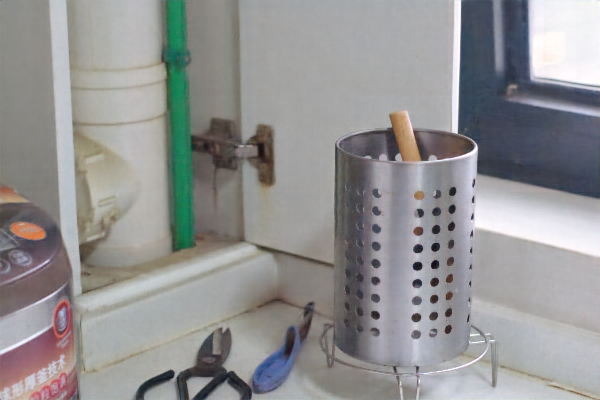}
     \centering
     \centerline{${\mathrm{BDCE}}$}
  \end{subfigure}
  \vspace{9pt}
  \caption{Visual results of ablation study. The naive DCE-based method can't denoise and obtain satisfactory result. $\mathrm{w/o}$ means without. Detail settings are provided in Sec.~\ref{Ablation Study} and Tab.~\ref{tableabs}}
  \label{fig:abs}
  \vspace{-10pt}
\end{figure*}

For the first problem, many LLIE methods~\cite{lore2017llnet,jiang2021enlightengan,yang2020fidelity,Wang_2019_CVPR,liu2021retinex,zhang2022deep,wu2022uretinex,yang2022adaint,xu2022snr,ma2022toward,kim2021representative,Wang_2022_ECCV,Yang_2022_ECCV} are not specifically designed for high resolution images, and these methods need to feed high resolution images into their networks, which is computationally expensive. The first problem can be solved using DCE-based methods~\cite{guo2020zero}, which achieve a small computational cost by downsampling the input image to a lower resolution, then predicting the Light-Enhancement curves 
 (LE-curves) at the low resolution. However, DCE-based methods~\cite{guo2020zero,zhang2021star} tend to use lightweight networks, resulting in poor curve estimation, and their pixel-wise adjustment causes them to fail to denoise, so they are less effective on real data as shown in Fig.~\ref{fig:abs} (naive result).

Hence, while the first problem can be addressed using the existing DCE-based methods, the second problem remains unresolved. Some existing diffusion models~\cite{chung2022diffusion,chung2022improving,chung2022parallel,wang2022zero} for image restoration also fail to solve these two problems, because they work on RGB pixel space, which makes them computationally expensive on high resolution images.

To deal with these two key problems, we proposes BDCE, an effective bootstrap diffusion model based high-resolution low-light enhancement and denoising method.
The main contributions are summarized as follows:
\vspace{-7pt}
\begin{itemize}
\setlength{\itemsep}{0pt}
\setlength{\parsep}{0pt}
\setlength{\parskip}{0pt}

\item In BDCE, a bootstrap diffusion model is presented for model the distribution of optimal curve parameters, which can then be used for high resolution images.
\item We set a denoising module into each iteration of curve adjustment for enhancement and denoising simultaneously.
\item Extensive experiments on benchmark datasets demonstrate the superiority of BDCE over previous state-of-the-art methods both quantitatively and qualitatively.
\end{itemize}

\section{Related Work}

\subsection{Learning-Based Methods in LLIE} 
Recently, there has been a notable surge in the development of deep learning solutions for addressing the LLIE problem~\cite{lore2017llnet,yang2022adaint,zhang2021star,zamir2020learning,Wang_2019_CVPR,wang2022ultra,wei2018deep,wu2022uretinex,zhang2019kindling,zhang2021beyond,liu2021retinex}. Wei \textit{et al.} \cite{wei2018deep} introduce a Retinex-based method that achieves superior enhancement performance in most cases while maintaining physical interpretability. DCC-Net \cite{zhang2022deep} employs a collaborative strategy based on a strategy of partition and resolve to preserve information of color and retain a natural appearance. Zhang \textit{et al.} \cite{zhang2019kindling} introduce KinD, which consists of three specialized subnetworks for layer decomposition and reflectance restoration, and illumination adjustment. LLNet \cite{lore2017llnet} utilizes a multi-phase encoder for reducing sparse noise to enhance and denoise images captured under low-light conditions. Wang \textit{et al.} \cite{wang2022ultra} present LLFormer, a computationally efficient approach that leverages blocks employing multi-head self-attention along different axes and cross-layer attention fusion for computational reduction. 
Wu \textit{et al.} \cite{wu2022uretinex} propose a novel approach where the Retinex decomposition problem is reformulated as a learnable network that incorporates implicit prior regularization. Liu \textit{et al}.~\cite{liu2021ruas} employ a cooperative prior architecture search strategy along with a principled optimization unrolling technique. Liang \textit{et al.} \cite{liang2022self} propose a unique self-supervised approach that optimizes a separate untrained network specifically for each test image. SCI \cite{ma2022toward} utilizes a self-supervised approach to autonomously adjust the reflection component. EnGAN \cite{jiang2021enlightengan} takes an unsupervised learning approach to tackle the challenges of overfitting and limited generalization. Zero-DCE~\cite{guo2020zero} performs pixel-level adjustments by leveraging a depth curve estimation network and a collection of non-reference loss functions. This combination allows for precise adjustments and improvements in image quality. Jin \textit{et al.}~\cite{jin2022unsupervised} propose a specialized network designed to suppress light effects and enhance illumination in darker regions. However, these methods are still challenging to solve the two problems mentioned in Sec.~\ref{intro}.
\vspace{-8pt}

\subsection{Diffusion Models}
    Within the domain of computer vision, diffusion models have garnered significant attention as a category of generative models. These models are trained to reverse the sequential corruption process of data by utilizing Gaussian random noise. Two main types of diffusion models have emerged:  score-matching based \cite{hyvarinen2005estimation,vincent2011connection} models and diffusion-based models \cite{sohl2015deep}. Notably, denoising diffusion probabilistic models \cite{ho2020denoising,nichol2021improved} and score networks \cite{song2019generative,song2020score,song2020improved,meng2021sdedit,chung2022come} conditioned on noise  have shown great promise in generating high-quality images. 
    In recent times, there has been a notable increase of interest in exploring the advantages of conditional forward processes in diffusion-based models. This exploration has demonstrated promising potential across diverse computer vision applications, including image synthesis \cite{dhariwal2021diffusion,kawar2022enhancing,ho2022classifier,zhang2023shiftddpms,chen2023fec,huang2023kv}, deblurring \cite{whang2022deblurring}, and image-to-image translation \cite{wang2022pretraining,saharia2022palette,choi2021ilvr}. Moreover, diffusion models have found applications in image restoration~\cite{huang2023wavedm}. For example, Ozan \textit{et al.} \cite{ozdenizci2022restoring} proposed a patch-based diffusion model for restoring images captured under challenging weather conditions. While many existing methods \cite{chung2022diffusion,chung2022improving,chung2022parallel,wang2022zero}  in image restoration focus on solving inverse problems and require prior knowledge of the degradation models, several concurrent works \cite{luo2023image,welker2022driftrec} have specifically addressed blind restoration problems such as deraining, deblurring, denoising, and face restoration. Kawar \textit{et al.} introduced DDRM \cite{kawar2022denoising} as a solution for linear inverse image restoration problems, but its applicability is limited to linear degradation. However, these diffusion based methods can't handle the first problems (high resolution image) mentioned in Sec.~\ref{intro}.

\section{Methodology}
    In this section, we describe the details of BDCE given a low-light image $\I_{l}$. Firstly, we describe the curve estimation for high resolution image in Section~\ref{sec:4.1}. Then, we design the bootstrap diffusion model for curve estimation in Section~\ref{sec:4.2}. Finally, we describe the denoising module for real low-light image in Section~\ref{sec:4.3}.

 \begin{figure}[t]
    \includegraphics[width=1\linewidth]{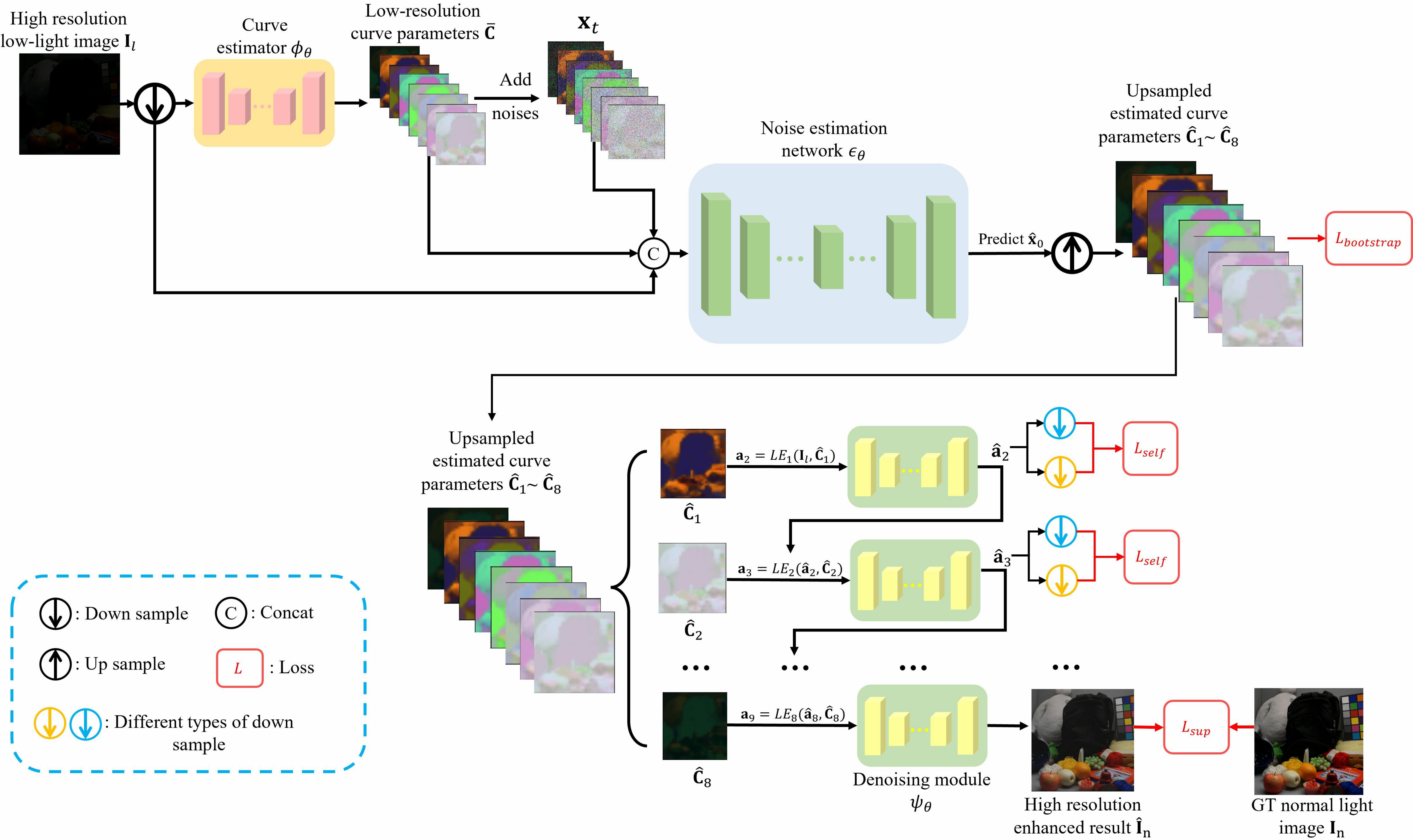}
  \vspace{-7pt}
  \caption{The training pipeline of our BDCE. We first downsample the high resolution low-light image and use curve estimator to predict the low-resolution curve parameters $\bar{\C}$. Then diffusion model is applied to learn a more accurate distribution of $\bar{\C}$. The estimated $\hat{\C}$ is upsampled and we adopt our denoise module in each iteration of curve adjustments to get a denoised final result.}
  \label{fig:pipe}
  \vspace{-10pt}
\end{figure}

\subsection{Curve Estimation for Hight Resolution Image}\label{sec:4.1}

Deep Curve Estimation Network (DCE-Net)~\cite{guo2020zero} is one of the most effective methods for enhancing low-light images since their adaptability to high resolution images.
In DCE-Net~\cite{guo2020zero}, a network is designed to predict a set of optimal Light-Enhancement curves (LE-curves) that match well with the input low-light image. This method maps all pixels of the input image by progressively applying curves to achieve the ultimate enhanced image. DCE-Net uses quadratic curves $LE(\I_l, \C)=\I_l+\C\I_l(1-\I_l)$ for mapping, where $\I_l \in \mathbb{R}^{3\times H \times W}$, $\C \in [-1,1]^{3\times H \times W}$ and $LE(\I_l, \C) \in \mathbb{R}^{3\times H \times W}$ denote the input image, the curve parameters and the adjusted image, respeatively. The curve parameters $\C$ are per-pixel predicted by a CNN in DCE-Net. 

The above procedure $LE(\I_l, \C)$ represents only one iteration of curve adjustment, in fact 8 curve adjustments are used in DCE-Net and each iteration can be denoted as 
\begin{equation}
    \y_{i+1}=LE_i(\y_i, \C_i)=\y_i+\C_i\y_i(1-\y_i),\quad i=1,...,8,\quad \y_1=\I_l.
\end{equation}
Then the total adjustment can be denoted by
\begin{equation}
    LE(\I_l, \C)=LE_8(\y_8, \C_8),\ \y_8 = LE_7(\y_7, \C_7),\ ...,\ \y_2 = LE_1(\y_1, \C_1).
\end{equation}
$\C_i \in [-1,1]^{3\times H \times W}$ and $\C = [\C_1,\ ...,\ \C_8] \in [-1,1]^{24\times H \times W}$. We express the above adjustment by $\C = \phi_\theta(\I_l)$ and $\I_{e} = LE(\I_l, \C)$, where $\phi_\theta$ is a CNN network.

Based on the localised nature of the curve adjustment, DCE-Net can efficiently cope with high resolution input images. For a high-resolution input image $\I_l$, DCE-Net first resizes to get a fixed low resolution image $\bar{\I}_l \in \mathbb{R}^{3\times \bar{H}\times \bar{W}}$ where $\bar{H}$ and $\bar{W}$ are always set to 256, then uses the curve estimation described above $\bar{\C} = \phi_\theta(\bar{\I}_l)$ to get a curve $\bar{\C} \in [-1,1]^{24\times \bar{H}\times \bar{W}}$ at the low resolution, then upsamples that curve $\bar{\C}$ to the original high resolution $\C \in [-1,1]^{24\times H\times W}$, and performs curve adjustment at high resolution. For any high resolution image, the computational cost of network is the same.

\subsection{Bootstrap Diffusion Model for Better Curve Estimation}\label{sec:4.2}
In order to make BDCE adaptable to arbitrary high resolution images, BDCE is very different from other diffusion model based methods, where other diffusion model based methods learns the distribution of target images, while our BDCE learns the distribution of curve parameters $\bar{\C} \in [-1,1]^{24\times \bar{H}\times \bar{W}}$. Because the target image $\I_n$ (the normal-light image in our task) may have a high resolution such that $\I_n \in \mathbb{R}^{3\times H\times W}$, the learning burden of $\I_n$ will be much bigger than it of $\bar{\C}$ if $H \gg \bar{H}$. 

The forward process in DDPMs aims to learn the distribution of $\x_0$, i.e., $\bar{\C}$ (curve parameters) in our BDCE. However, the values of $\bar{\C}$ are unknown at the beginning of training. To solve this, first we enter resized $\bar{\I}_l$ into curve estimator $\phi_\theta$ to get a curve estimation $\bar{\C}$.  Then, we denote $\x_0 = \bar{\C}$ as the data distribution of our diffusion model. Specifically, the forward process of BDCE can be described as follows:
    \vspace{-3pt}
    \begin{equation}\label{eqforward}
    q(\x_{t}\vert\x_0)=\N(\x_t;\sqrt{\bar{\alpha}_t}\x_0,(1-\bar{\alpha}_t)\I).
	\end{equation}
\begin{equation}\label{eqforwards}
    \x_{t}=\sqrt{\bar{\alpha}_t}\x_0+\sqrt{1-\bar{\alpha}_t}\epsilon_t,\ \epsilon_t \sim \N(\mathbf{0},\I)
	\end{equation}
With the same setting of $\Tilde{\mu}_t(\x_t,\x_0)$ in \cite{song2020denoising}, the reverse process from $\x_T$ to $\x_0$ is:
    \begin{equation}
    \begin{aligned}
		q(\x_{t-1}\vert\x_t,\x_0)=\N(\x_{t-1};\bm{\Tilde{\mu}}_t(\x_t,\x_0),\Tilde{\sigma}_t^2\I),
  \end{aligned}
		\label{DDIM sampling}
	\end{equation}
    where $\x_0$ can be predicted by a noise estimation network $\bm{\epsilon}_{\theta}(\x_t,\bar{I}_l,\bar{\C},t)$:
    \begin{equation}
	\begin{aligned}
		\boldmath{\hat{\x}}_0&=\frac{\x_t-\sqrt{1-\bar{\alpha}_t}\bm{\epsilon}_{\theta}(\x_t,\bar{I}_l,\bar{\C},t)}{\sqrt{\bar{\alpha}_t}}.
	\end{aligned}
	\label{DDIMx0}
	\end{equation}
 
As shown in Fig.~\ref{fig:pipe}, a U-Net similar to that in \cite{zagoruyko2016wide} is used as the noise estimation network $\bm{\epsilon}_\theta$ of BDCE. As for the curve estimator $\phi_\theta$, we adopt a lightweight network following~\cite{guo2020zero}. In each step $t$ of training, the given low-light image $I_l$ and the curve parameters $\bar{\C}$ serve as the conditions in BDCE to model the distribution of $\bar{\C}$. $L_{simple}$ defined in \cite{nair2022ddpm} is utilized as the supervision for $\bm{\epsilon}_{\theta}$. 

Note that the curve parameters $\bar{\C}$ estimated by $\phi_\theta$ are not the optimal curve parameters for enhancement. Therefore, we adopt a bootstrap diffusion model for stable training and better optimization for learning the distribution of $\bar{\C}$. To achieve this, the following bootstrap loss is used:
    \begin{equation}\label{bootstraploss}
    \begin{aligned}
        \mathcal{L}_{bootstrap}
        & = ||LE(\I_l, \hat{\C}) - \I_n||_2,\ \hat{\C} = \mathrm{Upsample}(\boldmath{\hat{\x}}_0), 
    \end{aligned}
    \end{equation}
    where $\hat{\C} \in [-1,1]^{24\times H\times W}$ denotes the estimated curve parameters from $\boldmath{\hat{\x}}_0$ in Eq. \ref{DDIMx0} upsampled to the original high resolution. 

 \subsection{Denoising Module for Real Low-Light Image}\label{sec:4.3}
	Because most of the real low-light images are degraded by real noises, LLIE always includes the demands of real image denosing. However, using only LE-curves is difficult to remove the real noises since curve adjustment is pixel-wise without localised smoothing, so all DCE-based methods suffer from the problem of real noises~\cite{guo2020zero}.
 \begin{figure*}[t]
  \scriptsize
  \centering
  \captionsetup[subfigure]{labelformat=empty}  
  \begin{subfigure}{0.118\linewidth}
    \includegraphics[width=1.45cm]{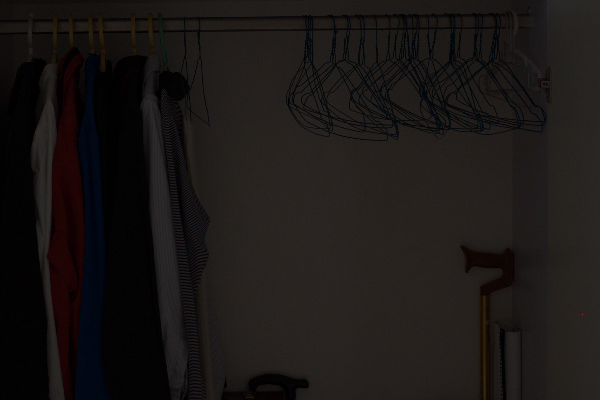}
     \centering
    \tabincell{c}{Input\\ \ }
  \end{subfigure}
  \begin{subfigure}{0.118\linewidth}
    \includegraphics[width=1.45cm]{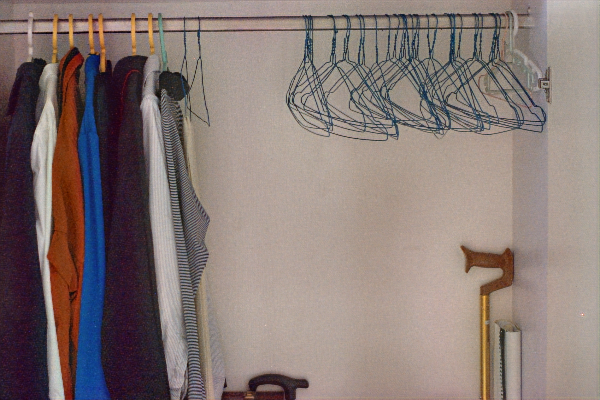}
     \centering
    \tabincell{c}{EnGAN\\ 23.28/0.804}
  \end{subfigure}
  \begin{subfigure}{0.118\linewidth}
    \includegraphics[width=1.45cm]{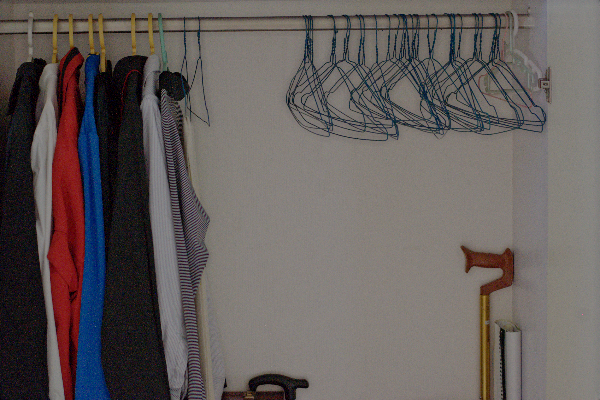}
     \centering
    \tabincell{c}{Zero-DCE\\ 15.45/0.7173}
  \end{subfigure}
  \begin{subfigure}{0.118\linewidth}
    \includegraphics[width=1.45cm]{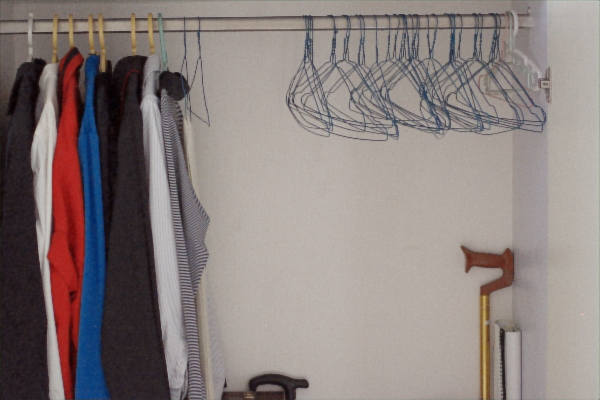}
    \centering
    \tabincell{c}{IAT\\ 26.49/0.8839}
  \end{subfigure}
  \begin{subfigure}{0.118\linewidth}
    \includegraphics[width=1.45cm]{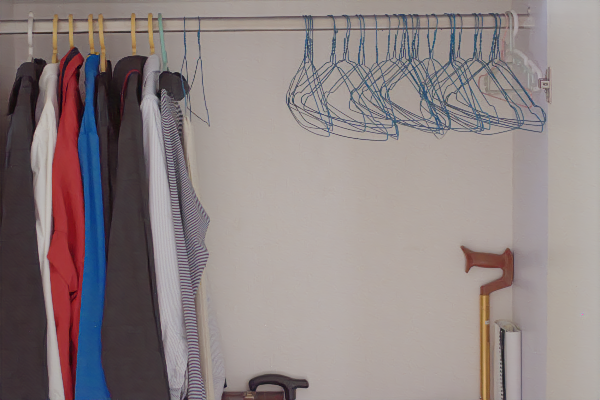}
    \centering
    \tabincell{c}{URetinex-Net\\ 26.85/0.912}
  \end{subfigure}
  \begin{subfigure}{0.118\linewidth}
    \includegraphics[width=1.45cm]{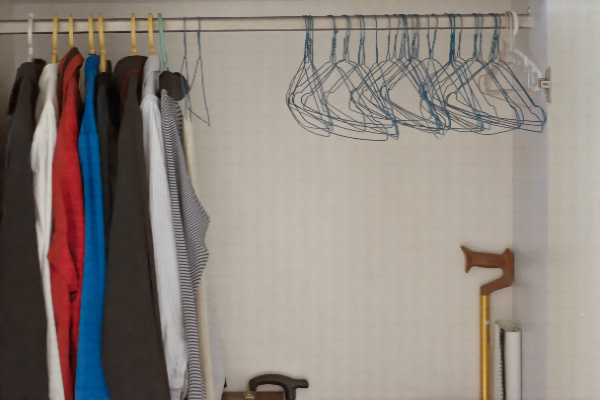}
    \centering
    \tabincell{c}{SNR \\ 27.58/0.904}
  \end{subfigure}
  \begin{subfigure}{0.118\linewidth}
    \includegraphics[width=1.45cm]{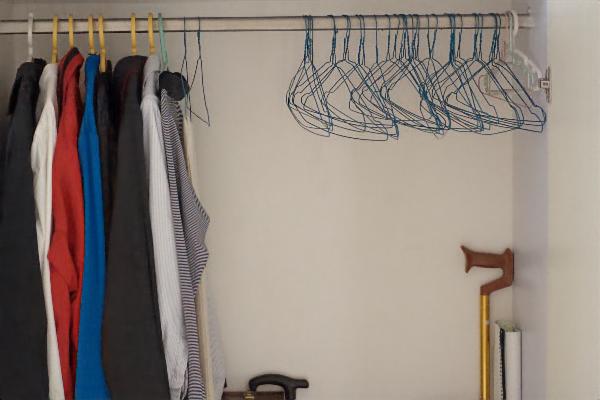}
    \centering
    \tabincell{c}{BDCE\\ 27.82/0.922}
  \end{subfigure}
  \begin{subfigure}{0.118\linewidth}
    \includegraphics[width=1.45cm]{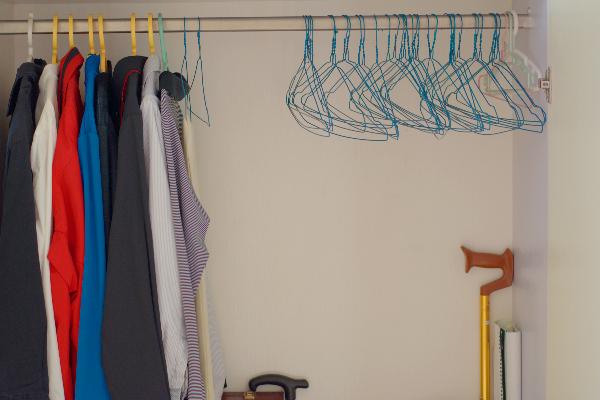}
    \centering
    \tabincell{c}{GT\\ PSNR/SSIM}
  \end{subfigure}
  \vspace{-5pt}
  \caption{Visual comparison on LOL-v1. BDCE yields less noise and more natural colors.}
  \label{fig:LOL}
  \vspace{-1pt}
\end{figure*}

\begin{figure*}[t]
  \scriptsize
  \setlength{\abovecaptionskip}{-0.3cm}
  \centering
  \captionsetup[subfigure]{labelformat=empty}  
  \begin{subfigure}{0.118\linewidth}
    \includegraphics[width=1.45cm]{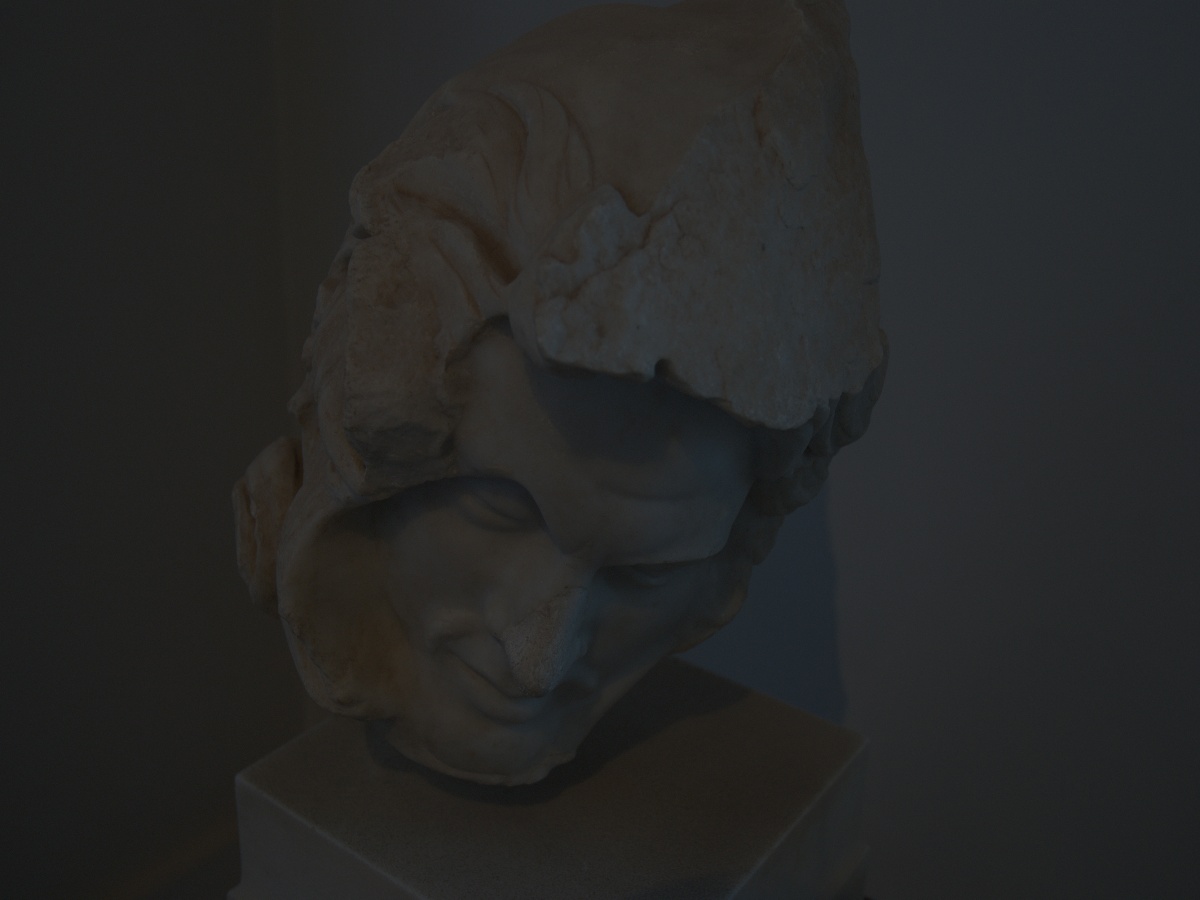}
    \centering
    \tabincell{c}{Input\\ \ }
    \label{fig:short-a}
  \end{subfigure}
  \begin{subfigure}{0.118\linewidth}
    \includegraphics[width=1.45cm]{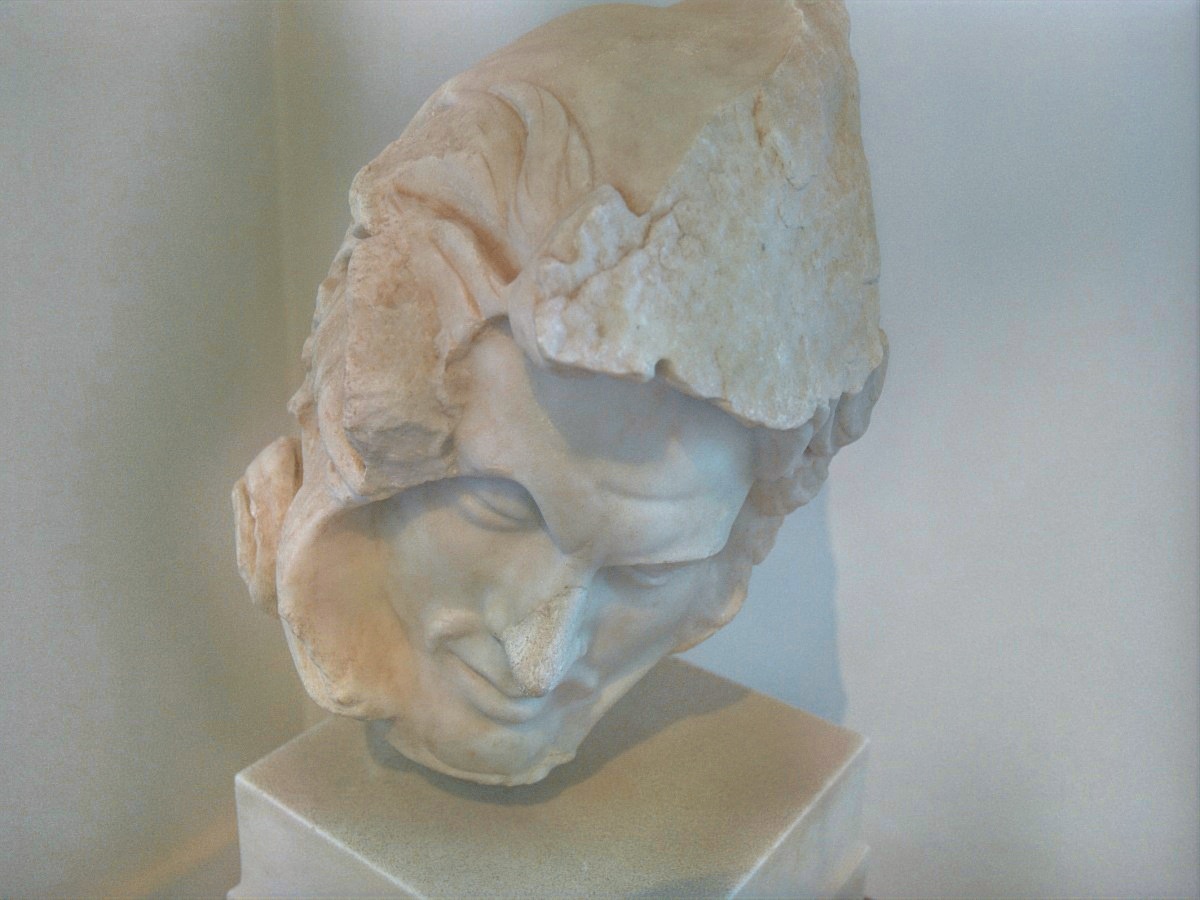}
    \centering
    \tabincell{c}{EnGAN \\ 12.89/0.736}
  \end{subfigure}
  \begin{subfigure}{0.118\linewidth}
    \includegraphics[width=1.45cm]{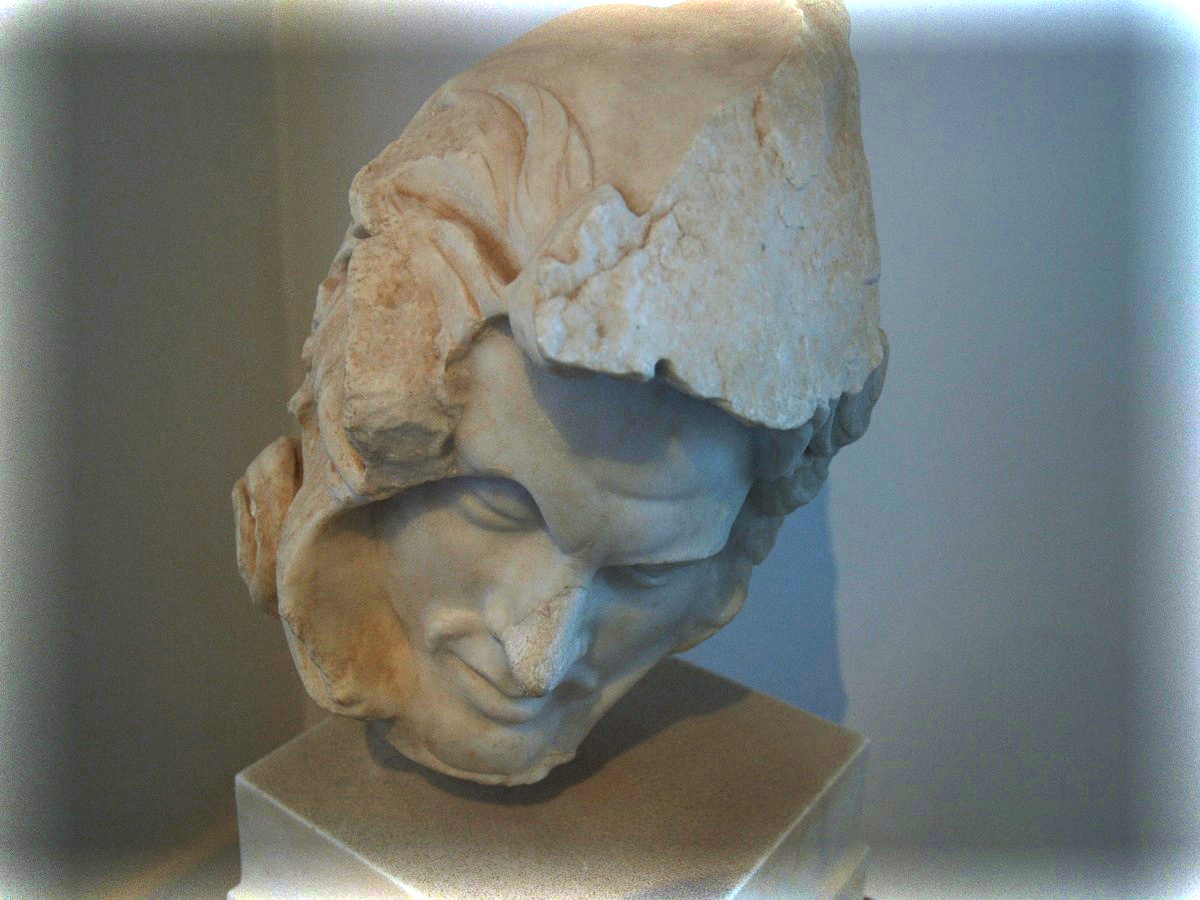}
    \centering
    \tabincell{c}{DeepUPE \\18.17/0.812}
  \end{subfigure}
  \begin{subfigure}{0.118\linewidth}
    \includegraphics[width=1.45cm]{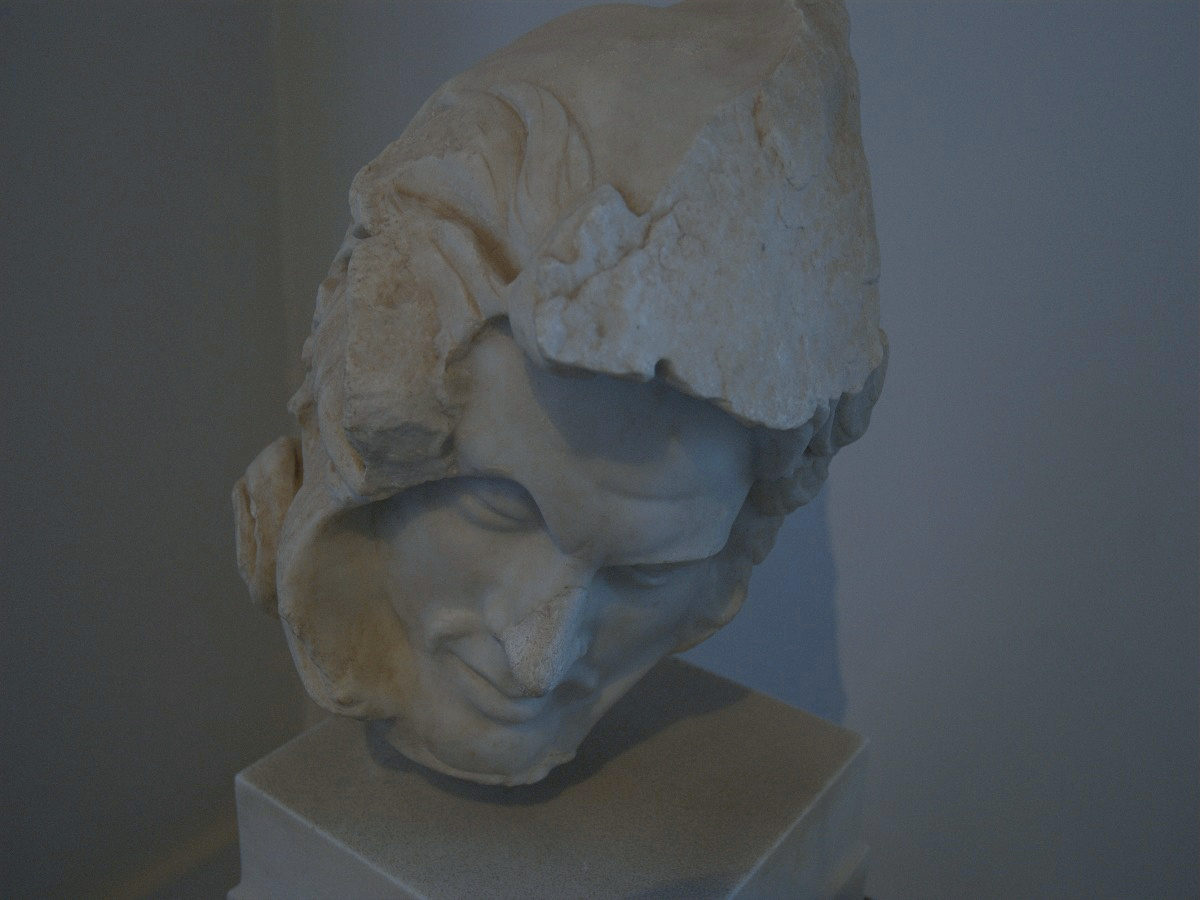}
    \centering
    \tabincell{c}{SCI\\15.99/0.766}
  \end{subfigure}
  \begin{subfigure}{0.118\linewidth}
    \includegraphics[width=1.45cm]{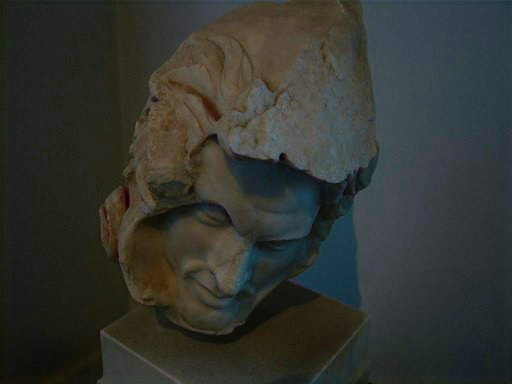}
    \centering
    \tabincell{c}{HWMNet\\12.43/0.584}
  \end{subfigure}
  \begin{subfigure}{0.118\linewidth}
    \includegraphics[width=1.45cm]{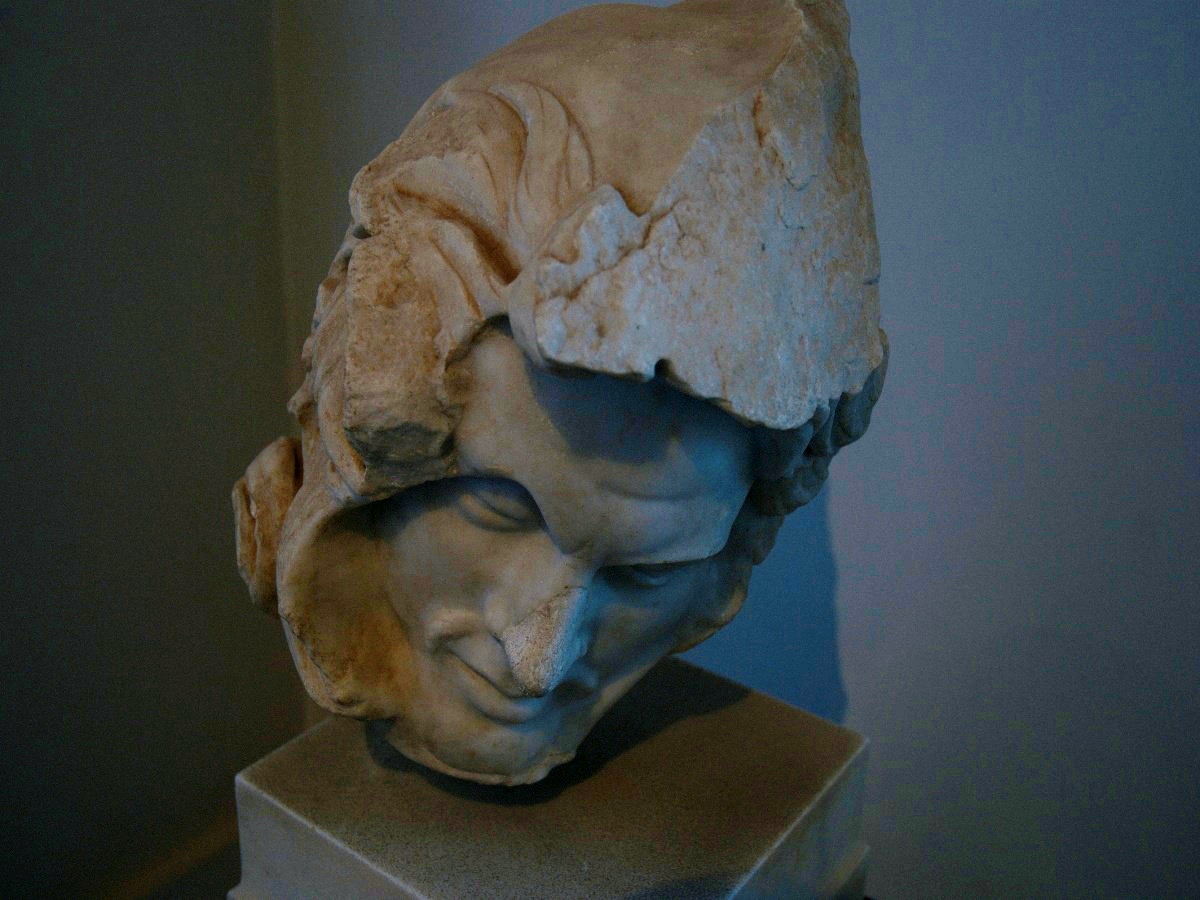}
    \centering
    \tabincell{c}{AdaInt\\15.49/0.763}
  \end{subfigure}
  \begin{subfigure}{0.118\linewidth}
    \includegraphics[width=1.45cm]{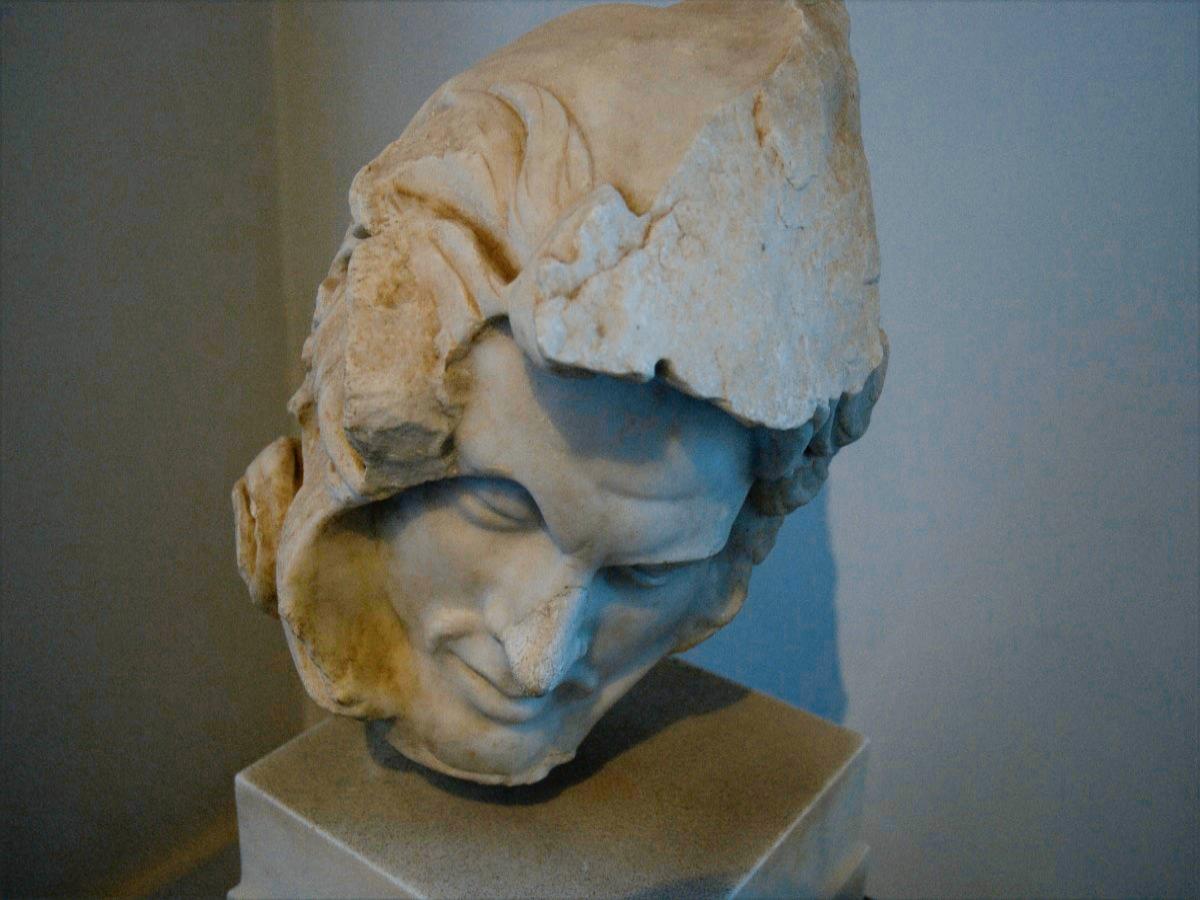}
    \centering
    \tabincell{c}{BDCE\\25.85/0.837}
  \end{subfigure}
  \begin{subfigure}{0.118\linewidth}
    \includegraphics[width=1.45cm]{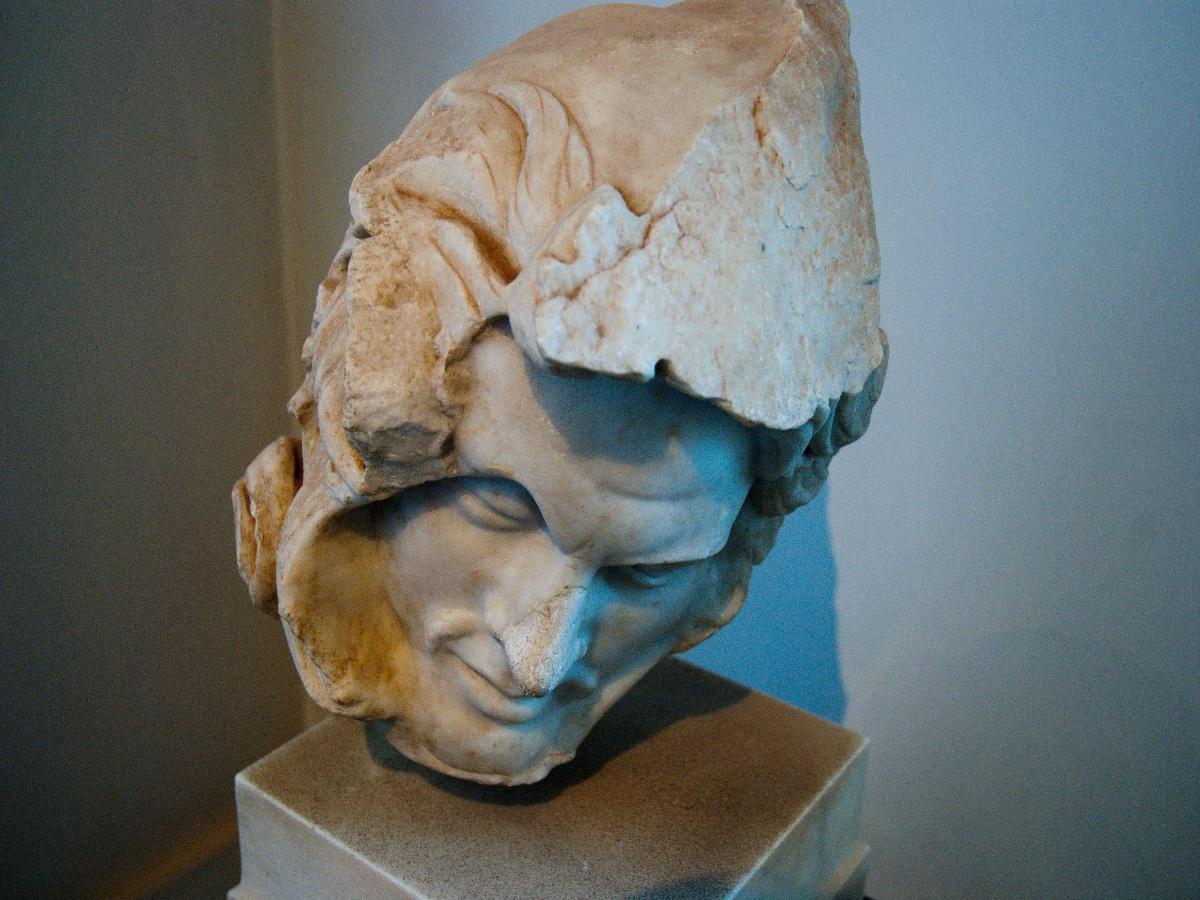}
    \centering
    \tabincell{c}{GT\\PSNR/SSIM }
  \end{subfigure}\\
  \vspace{13pt}
  \caption{Visual comparison on MIT-Adobe FiveK. BDCE yields better colors.}
  \label{fig:MIT}
  \vspace{-9pt}
\end{figure*}

    To solve this problem, we propose to combine enhancement and denoising in one model by applying a denoising module $\psi_\theta$ consisting of servel residul blocks. We implement the denoising during each iteration by refining the intermediate enhanced result $\y_i$ as following, given $\y_1=\I_l$:
\begin{equation}
\begin{aligned}
    &\y_2 = LE_1(\y_1, \hat{\C}_1),\ \hat{\y}_2 = \psi_\theta(\y_2),\\
    &\y_3 = LE_2(\hat{\y}_2, \hat{\C}_2),\ \hat{\y}_3 = \psi_\theta(\y_3),\\
    &...\\
    &\y_9 = LE_8(\hat{\y}_8, \hat{\C}_8),\ \hat{\I}_n = \psi_\theta(\y_9),
\end{aligned}
\end{equation}
where $\y_i$ denotes the intermediate enhanced result of the $i$-th iteration. The process described above can be briefly expressed as $\hat{\I}_n=LE_{de}(\I_l, \hat{\C})$. It can be understood as using the denoising module to denoise after each iteration of brightness adjustment before proceeding to the next iteration of brightness adjustment. After 8 iterations of brightness adjustment and denoising, the result is an enhanced output with less noise. To train $\psi_\theta$, a full-supervised loss $\mathcal{L}_{sup}$ and a self-supervised loss $\mathcal{L}_{self}$ are used:
\begin{equation}\label{denoiseloss}
\begin{aligned}
    &\mathcal{L}_{sup}
    = ||\hat{\I}_n - \I_n||_2,\\
    &\hat{\y}_i^{d1} = \mathrm{Down_1}(\hat{\y}_i),\ \ \hat{\y}_i^{d2} = \mathrm{Down_2}(\hat{\y}_i),\\
    & \mathcal{L}_{self}= \sum_{i=1, ..., 8}||\hat{\y}_i^{d1} - \hat{\y}_i^{d2}||_2,\\
\end{aligned}
\end{equation}
where $\mathrm{Down_1}$ and $\mathrm{Down_2}$ denote two types of downsampling way (randomly choose from max-pooling, average-pooling, nearest, bilinear, bicubic and lanczos interpolation), because the MSE between two different downsampling results $\hat{\y}_i^{d1}$ and $\hat{\y}_i^{d2}$ should be small for a denoised image.

\begin{figure*}[t]
\scriptsize
  \centering
  \setlength{\abovecaptionskip}{0.35cm}
  \captionsetup[subfigure]{labelformat=empty}  
  
  \begin{subfigure}{0.118\linewidth}
    \includegraphics[width=1.45cm]{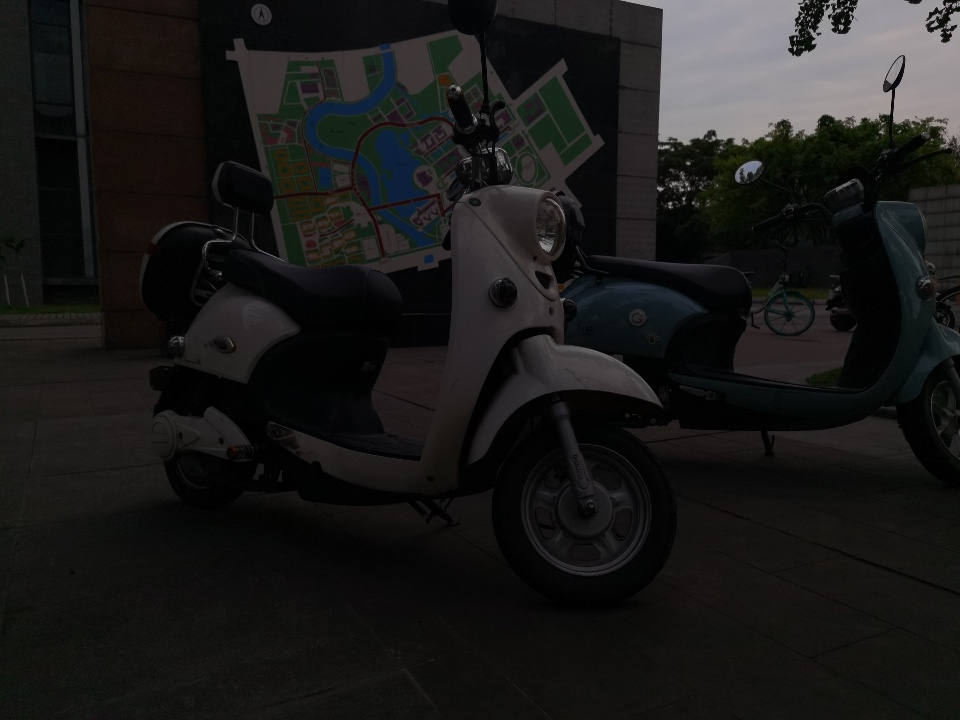}
    \centerline{Input}
  \end{subfigure}
  \begin{subfigure}{0.118\linewidth}
    \includegraphics[width=1.45cm]{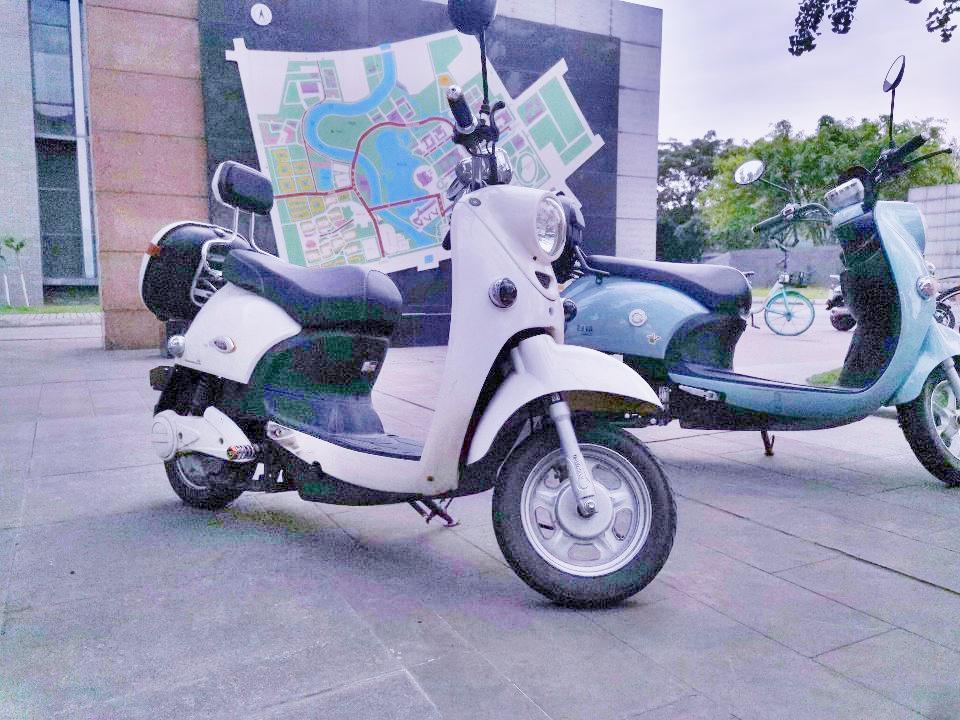}
    \centerline{Zero-DCE++}
  \end{subfigure}
  \begin{subfigure}{0.118\linewidth}
    \includegraphics[width=1.45cm]{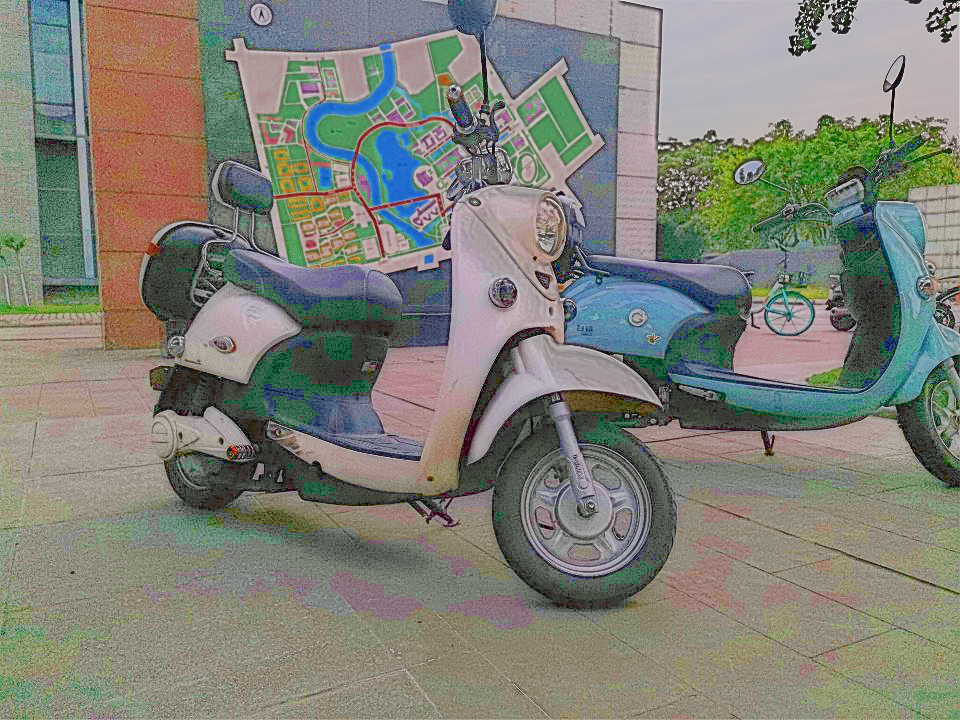}
    \centerline{Retinex-Net}
  \end{subfigure}
  \begin{subfigure}{0.118\linewidth}
    \includegraphics[width=1.45cm]{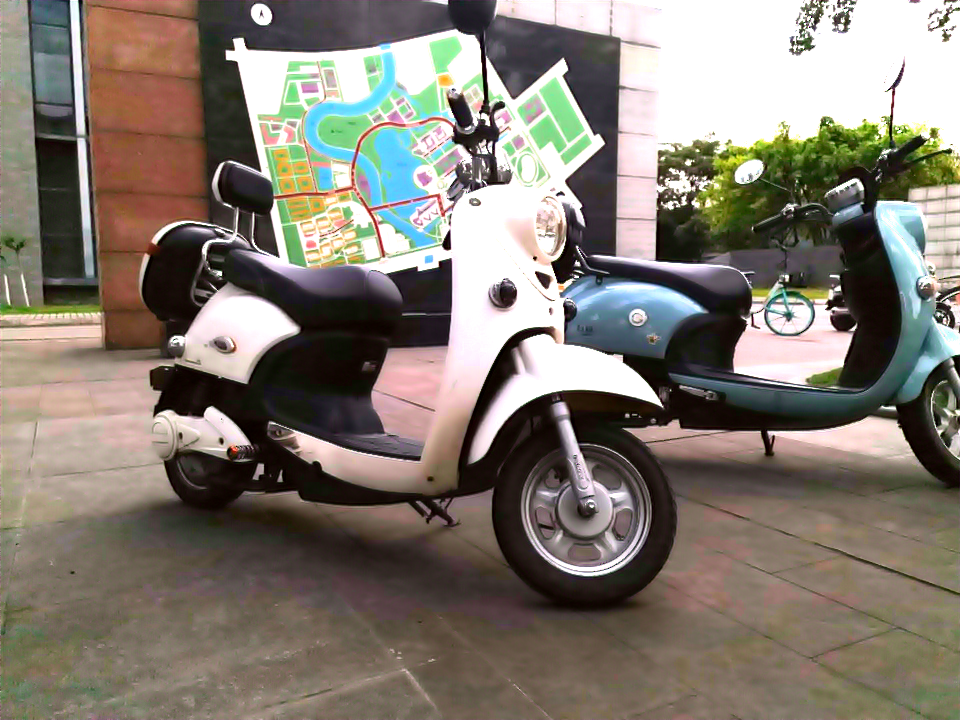}
    \centerline{RUAS}
  \end{subfigure}
  \begin{subfigure}{0.118\linewidth}
    \includegraphics[width=1.45cm]{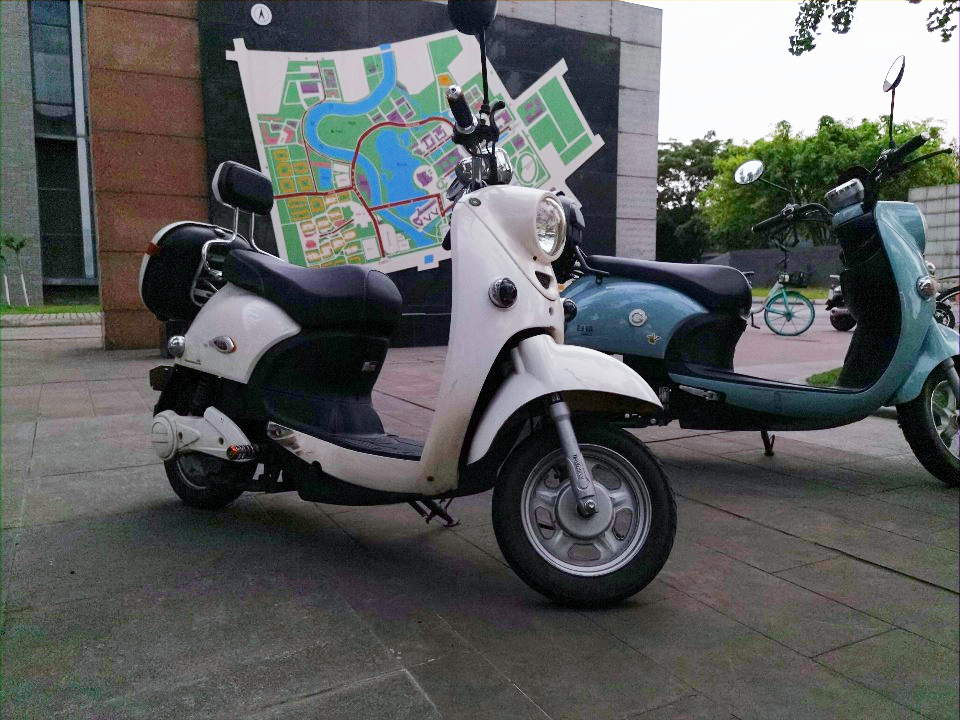}
    \centerline{SCI}
  \end{subfigure}
  \begin{subfigure}{0.118\linewidth}
    \includegraphics[width=1.45cm]{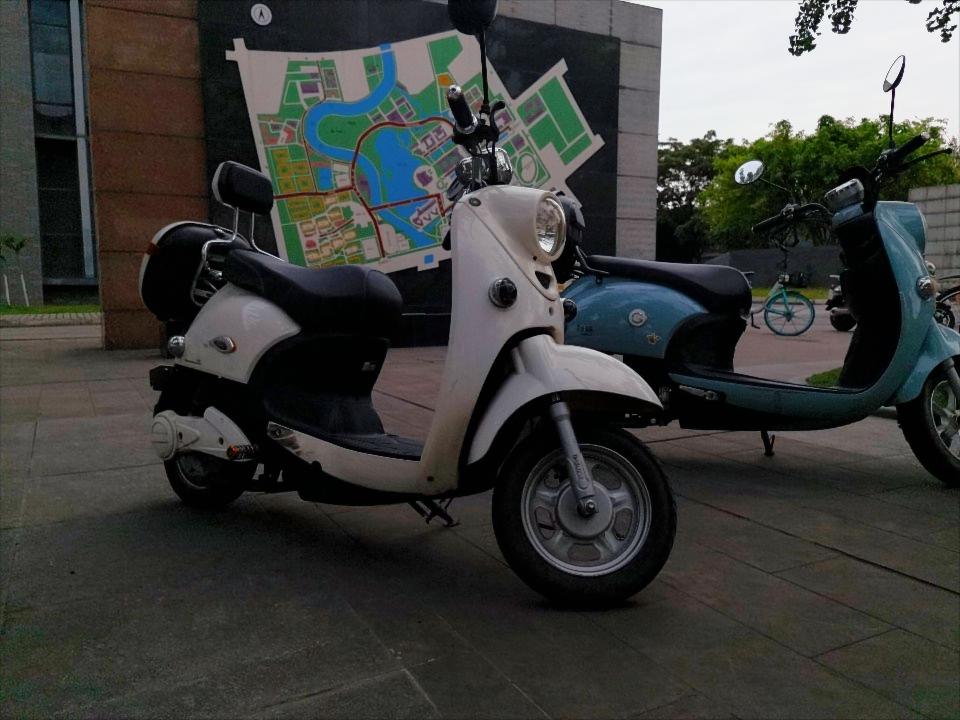}
    \centerline{SSIENet}
  \end{subfigure}
  \begin{subfigure}{0.118\linewidth}
    \includegraphics[width=1.45cm]{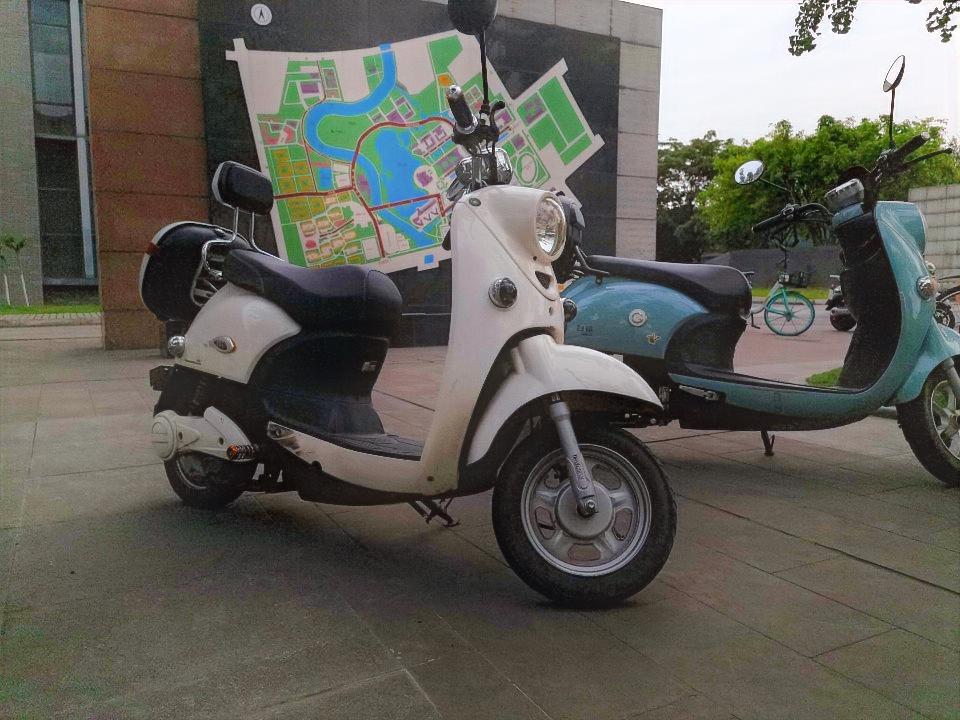}
    \centerline{BDCE}
  \end{subfigure}
  \begin{subfigure}{0.118\linewidth}
    \includegraphics[width=1.45cm]{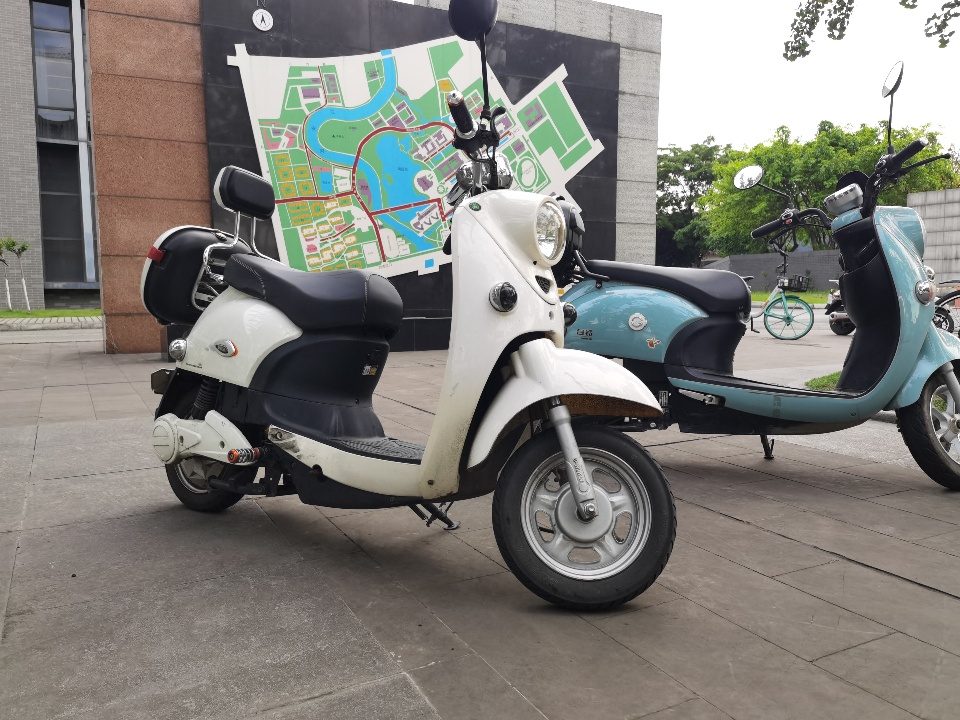}
    \centerline{Reference}
  \end{subfigure}
  \vspace{-9pt}
  \caption{Visual comparison on LSRW. BDCE yields natural colors with clear details.}
  \label{fig:LSRW}
  \vspace{-8pt}
\end{figure*}

\begin{table}[ht]
  \centering
  \scriptsize
  \setlength{\abovecaptionskip}{0.2cm}
  \setlength{\tabcolsep}{0.03mm}
  \resizebox{\linewidth}{!}{
    \begin{tabular}{ccccccccc}
    \toprule
    \multicolumn{9}{c}{LOL-v1}\\
    \toprule
    \makebox[0.06\textwidth][c]{Method} &KinD\cite{zhang2019kindling}  & Retinex\cite{wei2018deep}  &  DRBN\cite{yang2020fidelity} & URetinex\cite{wu2022uretinex} & IAT\cite{Illumination_Adaptive_Transformer} & SNR\cite{xu2022snr} & DCC-Net\cite{zhang2022deep} & BDCE  \\
    \midrule
    PSNR $\uparrow$ & 20.87 &18.23 & 19.55 &21.32 &23.38  &\underline{24.61} &22.72 &\textbf{25.01}\\
    SSIM $\uparrow$ & 0.800 &0.720  & 0.746 &0.834 &0.809 &\underline{0.842} &0.810 &\textbf{0.850} \\
    \bottomrule
    \multicolumn{9}{c}{LOL-v2-real}\\
    \toprule
    Method &KinD\cite{zhang2019kindling} & SID\cite{chen2018learning} &MIR-Net\cite{zamir2020learning} & A3DLUT\cite{wang2021real} &Retinex\cite{wei2018deep} & SNR\cite{xu2022snr} &Uformer\cite{wang2022uformer}& BDCE\\
    \midrule
    PSNR $\uparrow$ & 14.74 &13.24 &20.02 &18.19 &18.37 &\underline{21.48} &18.82 &\textbf{22.70} \\
    SSIM $\uparrow$ & 0.641 &0.442 &0.820 &0.745 &0.723 &\underline{0.849} &0.771 &\textbf{0.851} \\
    \bottomrule
    \multicolumn{9}{c}{LOL-v2-synthetic}\\
    \toprule
    Method  &KinD\cite{zhang2019kindling} & SID\cite{chen2018learning} &MIR-Net\cite{zamir2020learning} & A3DLUT\cite{wang2021real} & Retinex\cite{wei2018deep} & SNR\cite{xu2022snr} &Uformer\cite{wang2022uformer}& BDCE\\
    \midrule
    PSNR $\uparrow$  & 13.29 &15.04 &21.94 &18.92 &16.66&\underline{24.14} &19.66 &\textbf{24.93} \\
    SSIM $\uparrow$ & 0.578 &0.610 &0.876 &0.838 &0.652 &\textbf{0.928} &0.871 &\textbf{0.929} \\
    \bottomrule
    \multicolumn{9}{c}{MIT}\\
    \toprule
    Method & LCDPNet\cite{Wang_2022_ECCV} & DPE\cite{chen2018deep} &DeepUPE\cite{Wang_2019_CVPR}& MIRNet\cite{zamir2020learning} & HWMNet\cite{fan2022half} & STAR\cite{zhang2021star} & SCI\cite{ma2022toward}& BDCE\\
    \midrule
    PSNR $\uparrow$ & 23.23 &22.15 &23.04 &23.73 &\underline{24.44} &24.13 &20.44 &\textbf{24.85}\\
    SSIM $\uparrow$ & 0.842 &0.850 &0.893 &\textbf{0.925} &0.914 &0.885 &0.893 &\underline{0.911}\\
    \bottomrule
    \multicolumn{9}{c}{LSRW}\\
    \toprule
    Method&Retinex-Net\cite{wei2018deep}&KinD\cite{zhang2019kindling}&Zero-DCE\cite{guo2020zero}&SSIENet\cite{zhang2020self}&Zero-DCE++\cite{li2021learning}&RUAS\cite{liu2021ruas}&SCI\cite{ma2022toward}&BDCE\\
			\hline
			PSNR $\uparrow$ & 15.90& 16.47 &\underline{17.66}&16.74& 15.83 &14.43&15.01 &\textbf{20.10}\\
			SSIM $\uparrow$ & 0.3725 & \underline{0.4929} & 0.4685& 0.4879& 0.4664 & 0.4276& 0.4846 &\textbf{0.5308} \\
   \bottomrule
  \end{tabular}
  }
  \vspace{-3pt}
  \caption{Results on LOL-v1, LOL-v2, MIT and LSRW. The best and second best are in bold and underlined, respectively.}
  \label{tab:all}
  \vspace{-26pt}
\end{table}

\section{Experiments}

\subsection{Datasets Settings}
In our supervised experiments, we utilize several datasets, including LOL-v1 \cite{wei2018deep}, LOL-v2 \cite{yang2021sparse}, MIT-Adobe FiveK (MIT) \cite{bychkovsky2011learning} and LSRW \cite{hai2021r2rnet}. The LOL-v1 dataset consists of 485 training pairs and 15 testing pairs of real low-light images. As for LOL-v2, it consists of two parts: LOL-v2-synthetic and LOL-v2-real images. LOL-v2-synthetic contains 900 pairs of synthetic low-light images for training and 100 pairs for testing, while LOL-v2-real comprises 689 pairs of real low-light images for training and 100 pairs for testing. The MIT dataset comprises 5,000 paired synthetic low-light and normal-light images. We adopt the same training and testing settings as previous methods \cite{zhang2021star,Wang_2019_CVPR} for consistency. To assess the generalization ability of BDCE, we further evaluate its performance on unpaired real low-light datasets, including DICM \cite{lee2013contrast}, LIME \cite{guo2016lime}, MEF \cite{ma2015perceptual}, NPE \cite{wang2013naturalness}, and VV \cite{vv}.

\subsection{Comparison with SOTA Methods on Paired Data}
First, we assess the performance of our BDCE model through supervised training on paired datasets (LOL-v1, LOL-v2, MIT, LSRW). Subsequently, we evaluate the model's effectiveness by conducting testing on the corresponding test sets of these datasets.

\begin{table}[ht]
  \centering
  \resizebox{\linewidth}{!}{
    \begin{tabular}{@{}cccccccc@{}}
    \toprule
    Method &  Retinex-Net\cite{wei2018deep} & KinD\cite{zhang2019kindling}  & Zero-DCE\cite{guo2020zero} & EnGAN\cite{jiang2021enlightengan}& Zero-DCE++\cite{li2021learning}&SSIENet\cite{zhang2020self} &BDCE \\
    \midrule
    NIQE $\downarrow$ & 4.85 & \underline{4.09} &4.93 & 4.27&5.09&4.50&\textbf{3.99}   \\
    BRISQUE $\downarrow$ &27.77& 26.87&24.78&\underline{18.97}&21.11&21.08& \textbf{17.43}\\
    NIMA$\uparrow$ &\textbf{4.24}&\underline{4.16}& 3.87& 3.92&3.85&3.53&3.86 \\
    NRQM $\uparrow$ & \textbf{7.86} &7.47 &7.52 & 7.50 & 7.34& \underline{7.58}&7.44 \\
    PI $\downarrow$ & 3.15 &3.26 &3.21 &\underline{2.90} &3.29 & 3.47 & \textbf{2.87} \\
    \bottomrule
  \end{tabular}
  }
  \caption{Average results in terms of 5 NR-IQA metrics (NIQE, NIMA, BRISQUE, NRQM\cite{ma2017learning} and PI\cite{blau20182018}) on the 5 unpaired real low-light datasets (DICM, LIME, MEF, NPE and VV).}
  \label{tab:real5}
  \vspace{-17pt}
\end{table}

\begin{figure*}[ht]
   \scriptsize
  \centering
    \setlength{\abovecaptionskip}{0.1cm}
  \captionsetup[subfigure]{labelformat=empty}  
  \begin{subfigure}{0.118\linewidth}
    \includegraphics[width=1.45cm, height=1.12cm]{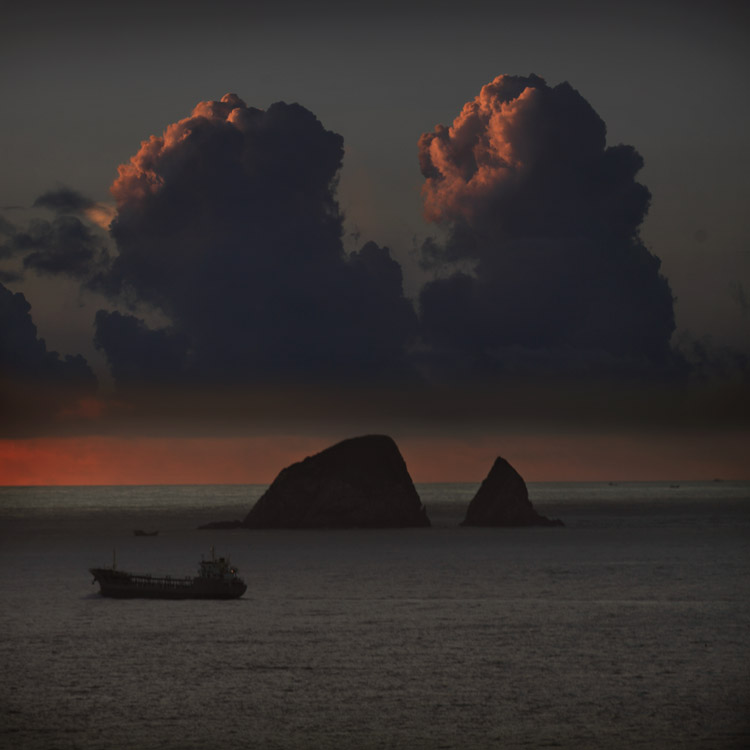}
    \centerline{Input}
  \end{subfigure}
  \begin{subfigure}{0.118\linewidth}
    \includegraphics[width=1.45cm, height=1.12cm]{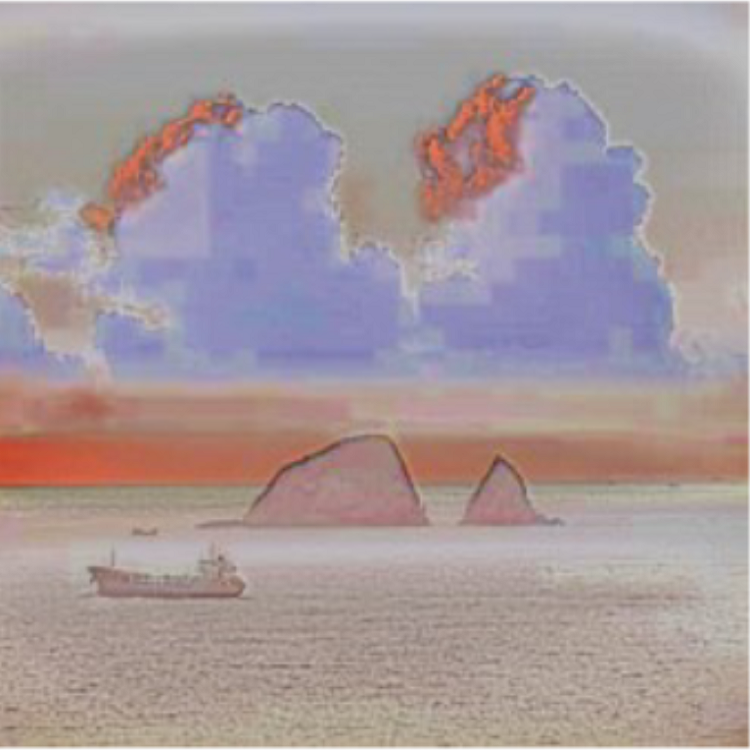}
    \centerline{Retinex-Net}
  \end{subfigure}
  \begin{subfigure}{0.118\linewidth}
    \includegraphics[width=1.45cm, height=1.12cm]{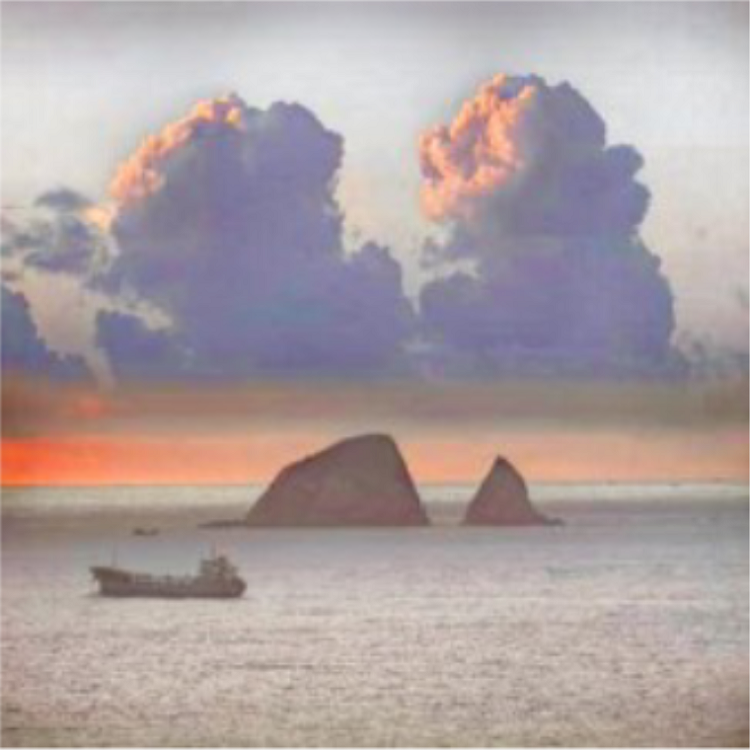}
    \centerline{EnGAN}
  \end{subfigure}
  \begin{subfigure}{0.118\linewidth}
    \includegraphics[width=1.45cm, height=1.12cm]{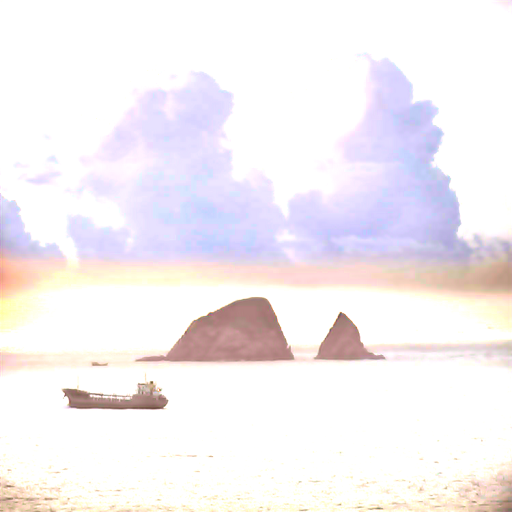}
    \centerline{RUAS}
  \end{subfigure}
  \begin{subfigure}{0.118\linewidth}
    \includegraphics[width=1.45cm, height=1.12cm]{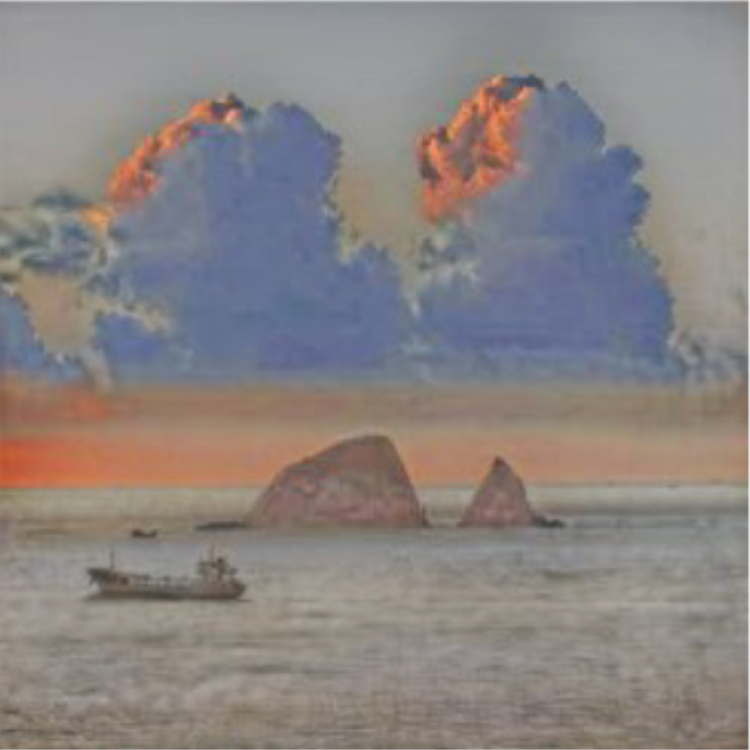}
    \centerline{KinD}
  \end{subfigure}
  \begin{subfigure}{0.118\linewidth}
    \includegraphics[width=1.45cm, height=1.12cm]{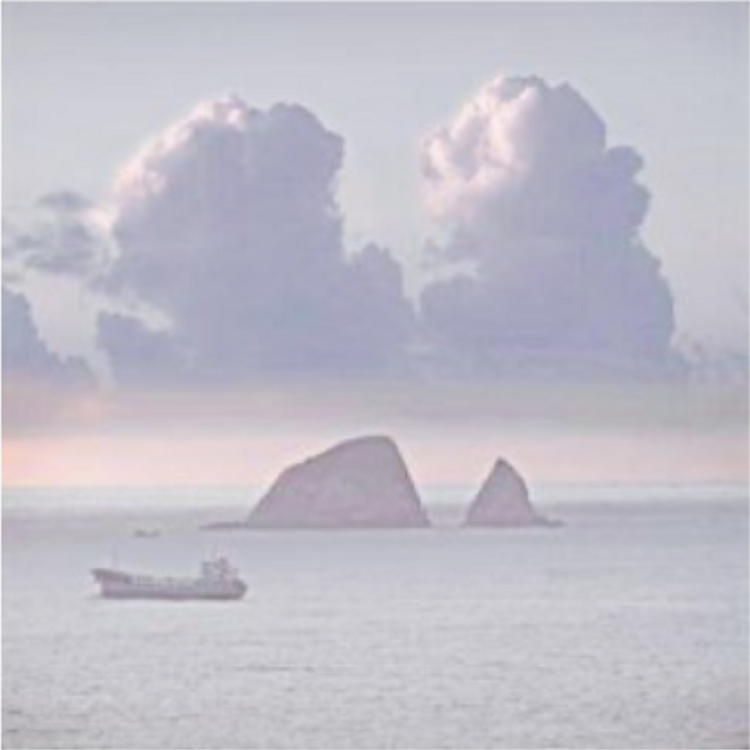}
    \centerline{Zero-DCE}
  \end{subfigure}
  \begin{subfigure}{0.118\linewidth}
    \includegraphics[width=1.45cm, height=1.12cm]{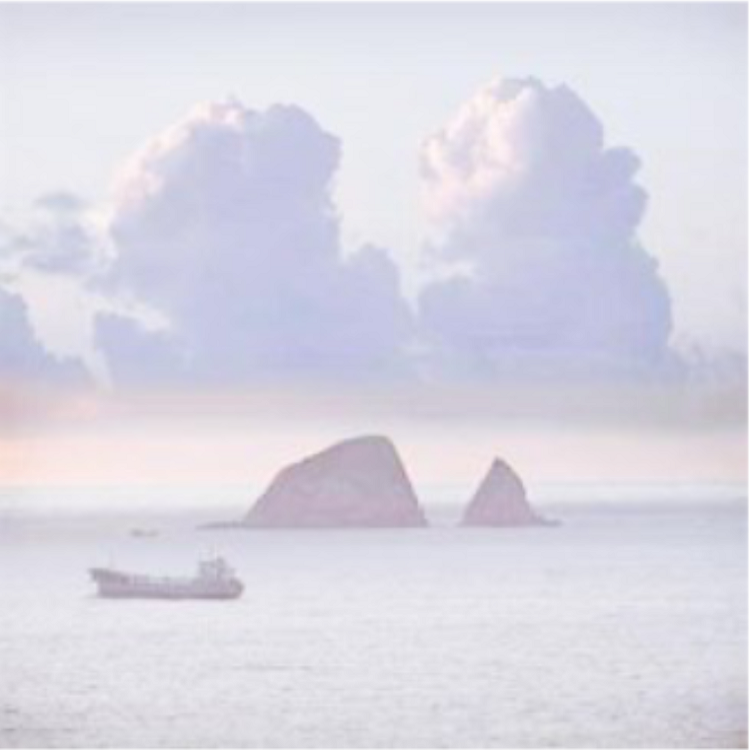}
    \centerline{Zero-DCE++}
  \end{subfigure}
  \begin{subfigure}{0.118\linewidth}
    \includegraphics[width=1.45cm, height=1.12cm]{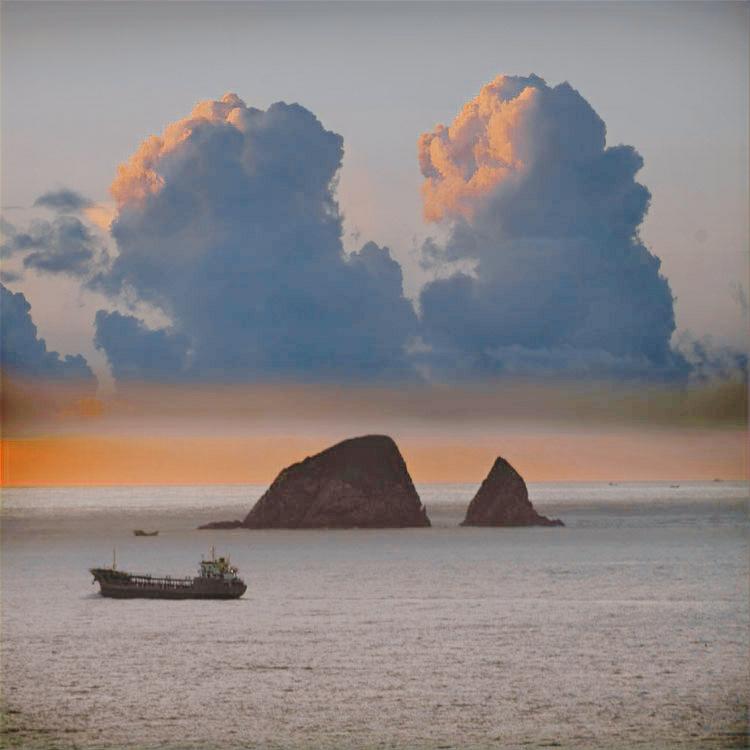}
    \centerline{BDCE}
  \end{subfigure}
  \vspace{-2pt}
  \caption{Visual comparisons on unpaired real low-light images, and the example is from the NPE dataset.}
  \label{fig:NPE}
  \vspace{-18pt}
\end{figure*}

\textbf{LOL Dataset.} Our BDCE is evaluated on LOL-v1 and LOL-v2 datasets. Results in the Table~\ref{tab:all} demonstrate BDCE's superiority over other state-of-the-art methods. The PSNR and SSIM metrics for the compared methods are sourced from their respective papers. Visual comparisons in the Fig.~\ref{fig:LOL} reveal that BDCE yields visually appealing results with reduced noise.

\textbf{MIT Dataset.} The performance of BDCE is assessed on the MIT dataset~\cite{bychkovsky2011learning}, and the obtained results in Table~\ref{tab:all} indicate that it achieves the highest PSNR and SSIM scores. Fig.~\ref{fig:MIT} clearly demonstrates that our BDCE method effectively prevents color-shift in the enhanced images. In contrast, some of the compared methods tend to exhibit over-enhancement or under-enhancement.

\textbf{LSRW Dataset}. Among the evaluated methods, BDCE achieves the highest scores on PSNR and SSIM, as shown in the Table~\ref{tab:all}. Fig.~\ref{fig:LSRW} presents a visual comparison of the results. While the images enhanced by other methods suffer from color shifts or appear under-enhanced, the images enhanced by BDCE exhibit a more natural appearance.
\vspace{-11pt}

\subsection{Comparison with SOTA Methods on Unpaired Data}

Our BDCE approach's effectiveness is evaluated on various unpaired datasets, namely DICM, LIME, MEF, NPE, and VV datasets. The evaluation is conducted by directly testing our pretrained model on the test set of each dataset. 

We compare the result of BDCE with other SOTA methods on these unpaired real low-light image datasets. The quantitative results in terms of 5 NR-IQA metrics are provided in a Table~\ref{tab:real5}. Additionally, a visual comparison is presented in Fig.~\ref{fig:NPE}, demonstrating that the results obtained by the compared methods often exhibit unrealistic appearances, loss of fine details, or excessive enhancement. In contrast, BDCE consistently produces images with enhanced colorfulness and sharp details.

\begin{table}[t]
		\small
		\begin{center}
			\setlength{\abovecaptionskip}{0.05cm}

			\setlength{\tabcolsep}{0.8mm}{
            \resizebox{\linewidth}{!}{
				\begin{tabular}{c|c|c|c|c|c}
					\toprule
					Method&Bootstrap Diffusion Model&Denoising Module&Self-supervised Loss&PSNR&SSIM\\
					\hline				${\mathrm{naive}}$&\XSolidBrush&\XSolidBrush&\XSolidBrush&18.51&0.721\\
                    ${\mathrm{w/o\  denoise}}$&\CheckmarkBold&\XSolidBrush&\XSolidBrush&22.15&0.809\\
                    ${\mathrm{w/o\  diff}}$&\XSolidBrush&\CheckmarkBold&\CheckmarkBold&23.33&0.807\\
                    ${\mathrm{w/o\ self}}$&\CheckmarkBold&\CheckmarkBold&\XSolidBrush&24.56&0.810\\
                    ${\mathrm{BDCE}}$&\CheckmarkBold&\CheckmarkBold&\CheckmarkBold&25.01&0.850\\
					\bottomrule 
			\end{tabular}}}
            \vspace{-1pt}
			\caption{Comparison of different settings in BDCE on LOL-v1. Bootstrap Diffusion Model: using the proposed bootstrap diffusion model for learning the distribution of curve parameters. Denoising Module: using the proposed denoising module in each iteration of curve adjustment. Self-supervised Loss: using self-supervised loss in Eq.~\ref{denoiseloss}. \CheckmarkBold: used. \XSolidBrush: not used.
			}\label{tableabs}
			\vspace{-35pt}
		\end{center}
	\end{table}
 
\subsection{Ablation Study}
\label{Ablation Study}

    We evaluate the performance of BDCE using various modules, presenting the results in Table~\ref{tableabs} and providing visual comparisons in Fig.~\ref{fig:abs}.
    
    The absence of the bootstrap diffusion model makes it challenging to acquire desirable curve parameters, leading to noticeable deficiencies in color rendition and illumination quality.
    
    Removing the denoising module results in severe noise in the enhanced output due to the inability of pixel-wise curve adjustment alone to effectively leverage the spatially local smooth prior for denoising.
    
    The denoising module's performance is enhanced by our self-supervised loss, which enables it to focus on denoising during each iteration of curve adjustment. Consequently, the utilization of the self-supervised loss proves advantageous. Overall, the combination of the proposed components effectively enhances the LLIE performance.

\vspace{-5pt}
\section{Conclusion and Limitation}
In this paper, we first analyse the problems of high computational cost in high resolution images and unsatisfactory performance in simultaneous enhancement and denoising. To mitigate these problems, we propose BDCE, a bootstrap diffusion model adapted to LLIE. For high resolution images, a curve estimation method is adopted and the curve parameters are estimated by our bootstrap diffusion model. At each iteration of curve adjustment, a denoise module is applied to denoise the intermediate enhanced result of each iteration. BDCE outperforms SOTA methods on LLIE benchmarks. 

The main limitation of BDCE is its time cost, which is due to the multiple steps of the sampling process of the diffusion model. For future research, we aim to devise a more streamlined approach for acquiring the curve parameter distribution. In addition, finding a lightweight network design is also a consideration.
%
%
\bibliographystyle{splncs04}
\bibliography{mybibliography}
%





\end{document}